\documentclass[preprint,12pt]{elsarticle}

\usepackage{amsfonts}
\usepackage{bbm}
\usepackage{xcolor}
\usepackage{amsmath,amssymb,amsfonts}
\usepackage{commath}
\usepackage{algorithm}
\usepackage{algorithmic}
\usepackage[ruled,vlined,linesnumbered, algo2e]{algorithm2e} 
\usepackage{natbib}
\usepackage{subcaption}
\usepackage{multirow}
\usepackage{url}
\usepackage{hyperref}
\usepackage{array}
\usepackage{booktabs}

\hypersetup{
	colorlinks=true,
	linkcolor=blue,
	citecolor=red,
}

\newcommand{\red}[1]{\textcolor{red}{#1}}
\newcommand{\scalemath}[2]{\scalebox{#1}{\mbox{\ensuremath{\displaystyle #2}}}}

\journal{Robotics and Autonomous Systems}

\begin{document}

\begin{frontmatter}



\title{Reinforcement Learning in Robotic Motion Planning by Combined Experience-based Planning and Self-Imitation Learning}


\author[a]{Sha Luo \corref{cor1}}
\ead{s.luo@rug.nl}
\author[a]{Lambert Schomaker}
\ead{l.r.b.schomaker@rug.nl}
\affiliation[a]{organization={University of Groningen},
            addressline={Nijenborgh 9}, 
            city={Groningen},
            postcode={9747 AG}, 
            country={The Netherlands}}

\cortext[cor1]{Corresponding author.}
            
\begin{abstract}
High-quality and representative data is essential for both Imitation Learning (IL)- and Reinforcement Learning (RL)-based motion planning tasks. For real robots, it is challenging to collect enough qualified data either as demonstrations for IL or experiences for RL due to safety consideration in environments with obstacles. We target this challenge by proposing the self-imitation learning by planning plus (SILP+) algorithm, which efficiently embeds experience-based planning into the learning architecture to mitigate the data-collection problem. The planner generates demonstrations based on successfully visited states from the current RL policy, and the policy improves by learning from these demonstrations. In this way, we relieve the demand for human expert operators to collect demonstrations required by IL and improve the RL performance as well. Various experimental results shows that SILP+ achieves better training efficiency, higher and more stable success rate in complex motion planning tasks compared to several other methods. Extensive tests on physical robots illustrate the effectiveness of SILP+ in a physical setting. 
\end{abstract}




\end{frontmatter}

\section{Introduction}
Motion planning is a fundamental module employed in many robotic platforms \cite{4651222,yao2021singularity, 8643443,rosenbaum2001posture,8794317,4543471}. For manipulators, sampling-based motion planning (SBMP) methods, including Rapidly-exploring Random Tree (RRT) \cite{lavalle1998rapidly}, Probabilistic Roadmap (PRM) \cite{kavraki1996probabilistic}, are widely used in recent decades. These methods can be easily implemented on robots with high-dimensional degrees of freedom (DoF) as they approximate the collision-free space by sampling, instead of depending on the explicit geometry modeling of the collision and collision-free configuration spaces. In addition, they are probabilistically complete. However, the main disadvantage of SBMP is the slow adaptability in dynamic environments. In traditional trajectory planning, the assumption is that all necessary task-space information is available at the start of planning and that the world does not change when movement starts. Driven by the growing demand for intelligent autonomous robots that react instantly in changing environments, there is an increasing interest in neural motion planners (NMPs)~\cite{Tamar-RSS-19,8793889, ravichandar2020recent} which approximate the planners with neural networks, such as imitation learning (IL) and reinforcement learning (RL). 
Instead of relying on a precise pre-planned path, such systems can use a more general policy for dealing with, e.g., unexpected obstacles, in order to adapt motion control, online.

Data collection is the primary barrier for training NMPs in high-DoF manipulation tasks due to the requirement of massive, diverse data in neural-based IL and RL. However, for IL, the main challenge is the lack of representative data near the boundary of obstacles when the objective is to learn obstacles-avoiding behaviors~\cite{Tamar-RSS-19}. For RL, exploring regions around obstacles to collect training data is unsafe and impractical in the real world. When there is inadequate or unbalanced data, it takes a long time for RL algorithms to gather enough informative experience for training the policy such that the algorithms perform well in a dynamic environment. The combination of IL and RL can potentially boost the performance in IL due to the improved exploration and it can accelerate the convergence in RL with the exploitation of "expert" knowledge. However, a heavy data-preparation process still needs to be realized to bootstrap the approach. This can be realized using non-human examples from a simulated planner~\cite{6174325}, via demonstrations by a human user~\cite{Rajeswaran-RSS-18,sauser2012iterative,kober2011policy} or by gathering data in parallel from multiple robots~\cite{kalashnikov2018qt} in an ensemble. 

Considering the advantages and disadvantages of traditional and learning-based motion planning algorithms, we explore the integration of planning and learning in motion planning tasks to tackle the data collection problem. On the learning side, RL makes decisions on selecting the next move in order to reach the goal configuration. As we know, RL is a process that balances exploration and exploitation. However, most experiences in the early training stage are not being exploited efficiently when they are not involved in successful trials. Still, such experiences could facilitate the obstacle-avoiding behaviors with collision-free biases. Therefore, we regard these episode-explored states as {\em candidate collision-free nodes} for the graph-based motion planning algorithms. In this work, we use a PRM-based method for supportive, online demonstration generation during training. By planning on the {\em candidate collision-free nodes}, this method generates up-to-date demonstrations, episode by episode, and saves them in the demonstration experience replay buffer. Since the demonstrations are generated based on the experience from the RL policy, we regard the planning module as a form of \emph{experience-based planning}. We exploit those demonstrations for imitation learning, which is similar to the idea of learning from good past experiences. We categorize this method as a self-imitation learning algorithm and call it {\em self-imitation learning by planning}. The self-imitation learning scheme guides the RL to learn from demonstrations automatically and continuously without requiring a human expert for laborious data collection.

In this paper, we propose a new algorithm, called SILP+, which is an enhanced version of SILP \cite{9561411}. Compared to SILP, SILP+:
\begin{enumerate}[1)]
    \item proposes a Gaussian-process-guided exploration method to reduce undesirable collisions near obstacles. The reduced number of collisions improves the safety during exploration process and reduces the training time by 18\%. The importance and details are explained in section 3.5, and experimental results can be check in section 4.4.
    \item analyzes the extrapolation error in actor-critic neural networks and proposes a reward-based filter to stabilize the training process and helps increase the success rate. Motivations and differences are explained section 3.4, and the experimental results can be check in section 4.5.
    \item provides detailed and extensive analysis on different methods of dealing with collision failures during training. We discover that the RL agent learns better with positive feedback, but negative experience also helps. Details and discussions can refer to section 4.3.
\end{enumerate}

Besides, we tested SILP+ in a pick-and-place task with the physical UR5e platform. The experimental results verify the robustness and adaptability of SILP+ in noisy and uncertain environments. 

Although the current study mainly focuses on motion-planning tasks for manipulators, the empirical findings and analysis are of interest to the general study of motion planning and reinforcement learning due to the following reasons: First, SILP+ presented here is an algorithm that combines planning and learning to improve the performance of motion planning. It represents a new way of utilizing the advantages of planning in a learning scheme, while avoiding a high computation load; Second, the empirical results, the analysis of the collision failures and the analysis of the extrapolation error in actor-critic networks provide helpful inspiration for designing high-performance RL architectures for motion-planning tasks.   

The remainder of this paper is organized as follows. Section \ref{RelatedWork} introduces the background and related work about neural motion planners and learning from demonstration, followed by the proposed methodology SILP+ in Section \ref{Methodology}. In Section \ref{Experiments}, experiments are conducted in both simulations and on the physical robot arm UR5e to analyze the efficiency and feasibility of the proposed SILP+ in motion planning tasks. Section \ref{Conclusion} draws conclusions and envisions future work.

\section{Related Work}\label{RelatedWork}
\subsection{Neural motion planning}
Methods using neural networks to assist or approximate motion planners have attracted increasing attention because of recent advances in deep learning and the demand for adaptable motion planners in changing environments. Recent work includes biasing sampling to critic regions with machine learning techniques. For example, Zhang et al. \cite{8594028} employed RL to learn the probability of rejecting a sampled state in order to bias the sampling to the less dangerous regions. Ichter et al. \cite{ichter2018learning} predicted sampling nodes in critic regions on top of traditional SBMP methods using a conditional variational autoencoder. Francis et al. \cite{francis2020long} proposed PRM-RL for long-range navigation, in which RL functions as a short-range local planner and also a collision-prediction module for the high-level PRM planner. A similar framework can be seen in \cite{angulo2022policy} which uses a regular global planner (RRT or A*) and an RL policy optimization method as the local planner. Besides, there are pure neural motion planners that approximate the motion planners with neural networks, mapping states directly to paths or actions. Qureshi et al. \cite{8793889} proposed MPNet as the neural motion planner, and the MPNet showed less planning time compared with traditional motion planners. Jurgenson~et~al.~\cite{Tamar-RSS-19} adapted DDPG and learned collision and reward models for visual motion planning tasks, which improved the accuracy and planning time compared with SBMP methods. The combination of planning with RL has also been investigated. Benjamin~et~al.~\cite{eysenbach2019search} proposed a hierarchical framework for long-horizon reaching tasks: planning at the high level to find the subgoals in the replay buffer and learning at the low level to control the robot to reach the subgoals. It demonstrated how efficient it was when planning was embedded with learning. However, they focused on 2D reaching tasks; no obstacles existed in the environment and they planned on the whole replay buffer, which may involve heavy computation in a more complex task. Differently, Xia~et~al.~\cite{9561315} used a SBMP planner at the low level, planning from the current state to subgoals and training the off-policy RL algorithm at the high level for subgoals generation. 

\subsection{Learning from Demonstrations}
Learning a task from scratch without prior knowledge is a daunting process; even human beings and animals rarely try to learn from scratch \cite{schaal1996learning}. They utilize previous experiences and demonstrations from instructors to derive strategies to approach a learning problem, which is called learning from demonstrations (LfD) or IL. LfD is widely used in learning-based robotic tasks, including helicopter maneuvering \cite{ng2003autonomous} \cite{abbeel2007application}, mobile robot navigation \cite{8976128}\cite{9117021}, surgery \cite{7487167}\cite{9196588}, manipulation \cite{8463162} \cite{4813845}. However, there are also limitations in LfD caused by sparse or poor datasets. Firstly, the controlling error will accumulate in behavior cloning when the agents encounter unfamiliar and unseen states in the demonstrations. Secondly, the policy's performance depends heavily on the demonstrations' quality; the agent cannot perform better than the supervisor without additional information to help improve. \cite{atkeson1997robot} \cite{ross2010efficient}. 

One of the solutions is the combination of RL and LfD, called reinforcement learning from demonstrations (RLfD) \cite{brys2015reinforcement} \cite{jing2020reinforcement}, which exploits the strengths of both sides and overcomes their shortcomings. The demonstrations are used to guide and improve RL policies, and then the RL provides feedback on the actions via the reward function and explores better policy than that of the supervisor~\cite{vecerik2017leveraging,hester2018deep,8463162} by exploration. Our SILP+ is similar to the work presented in~\cite{8463162}, where demonstrations were stored in a demonstration replay buffer and embedded in an auxiliary behavior cloning loss to guide the learning. We modify the LfD framework by adding a planning module for demonstration generation, and these demonstrations are utilized for further self-imitation learning. Similar to SILP+, DAgger in \cite{ross2011reduction} also adopts the idea of online supervision, which gives on-time evaluation feedback for the encountered states by relabelling actions. However, unlike DAgger, which retrieves expert guidance on every singe step, SILP+ plans on {\em all} states experienced within an episode, and therefore, it discovers the core steps in the episode through the global knowledge.

\subsection{Self-Imitation Learning}
The main idea of Self-Imitation Learning (SIL)~\cite{oh2018self} is to improve the sample efficiency in RL by utilizing good decisions in the past. However, the quality of the method depends heavily on the RL exploration strategies. It is difficult for the RL policy to obtain informative steps without direct supervision in complex robotic tasks. Besides, SIL was initially designed for the on-policy discrete settings, which is not straightforward to be used in off-policy, continuous action scenarios. Recent work \cite{dai2020episodic} targeted at the robotics applications with continuous action space and proposed ESIL that combined hindsight with SIL such that the agent learned from good experience selected by Hindsight Experience Replay (HER) \cite{andrychowicz2017hindsight}. The major difference between ESIL and SILP+ is how they create good experiences. ESIL changed the goal based on HER to transform useless experience to positive feedback, while SILP+ uses rigid planners to convert ordinary trajectories into optimized successful paths. Therefore, SILP+ collects higher-quality experiences for SIL to learn.

\subsection{Leveraging Prior Experience}
SILP+ can also be categorized as a data augmentation method, manipulating previous experience for better training. In this category, Weber et al. \cite{racaniere2017imagination} learned the policy with data from imaginations, in which part of the training data was aggregated by rolling out the policy. The widely known HER \cite{andrychowicz2017hindsight} is also a method of utilizing prior experience. It generates informative data by regarding the experienced next state as the goal. However, the amount of useless data increases as more successful experiences are gained in HER, and thus impairs the sample efficiency. Nevertheless, Our method can select the most promising state as the next state, and those selected states in the episode could form a successful path to provide more informative data for RL training.

\section{Methodology}\label{Methodology}
The combination of RL and LfD has been used to tackle the problem of sample efficiency in RL. It is straightforward and efficient, but the preparation of expert demonstrations might not be easy in many situations. Therefore, a more practical approach is learning by utilizing the agent's informative experiences. In the context of RL, SIL is a technique that takes advantage of this idea and encourages the agent to learn from actions with higher rewards. However, this technique was proposed for discrete control tasks and faced the problem of lacking informative, positive experiences in high-dimensional continuous tasks, such as robotics motion control.

Our SILP+ is a combination of the concepts of SIL and experience-based planning. Experience-based planning ensures qualified guidance by providing direct corrections on visited states using planning methods, while SIL guides the policy learning with good examples from the planning module. This section first formulates the motion planning task in an RL scheme, followed by the explanation of experience-based planning with PRM. Then, we introduce SIL in continuous control tasks and illustrate how planning is embedded with SIL to formulate SILP+. In addition, we propose a Gaussian-process-guided exploration method near collision regions to improve the exploration efficiency. Finally, a model-based reward filter is employed to reduce the extrapolation error in actor-critic RL.

\subsection{RL for Motion Planning}
We formulate our motion planning task as follows. Let $\mathcal{W}$ represent the world space. The set of obstacles in the world space is denoted by $\mathcal{O}$. The configuration space (i.e., the $\mathcal{C}$-space) is denoted by $\mathcal{C}$, in which a configuration of the robot is denoted by $q\in \mathcal{C}$. The forward kinematics $\mathcal{FK}: \mathcal{C} \to \mathcal{W}$ maps the robot configuration in $\mathcal{C}$ to the world space $\mathcal{W}$. If $\mathcal{FK}(q)$ in $\mathcal{W}$ belongs to the obstacle set $\mathcal{O}$, then $q$ is in the collision configuration space $\mathcal{C}_{obs}$. The collision-free configuration space then is defined by: $\mathcal{C}_{free} = \mathcal{C} \setminus \mathcal{C}_{obs}$. Given the collision-free starting configuration $q_0 \in \mathcal{C}_{free}$ and the goal configuration $q_g \in \mathcal{C}_{free}$, the motion planning task is to find a path that starts from $q_0$ and ends at $q_g$, while avoiding the collision configuration space $\mathcal{C}_{obs}$. This task can be formulated as a Markov Decision Process~(MDP), which can be solved in the reinforcement learning framework. 

We embed our motion planning task in an episodic off-policy RL framework, in which the environment and task are randomly sampled within the workspace at each episode \cite{sutton2018reinforcement}. The goal of the agent is to maximize the expected accumulated future returned reward $R_{t}=\mathbb{E}[\sum_{i=t}^\infty\gamma^{i-t}r_{i+1}]$ from the current step $t$ with a discounted factor $\gamma \in [0, 1]$ weighting the future importance. Each policy $\pi$ has a corresponding action-value function $Q^{\pi}(s, a)=\mathbb{E}[R_t|s_t=s, a_t=a]$, representing the expected return under policy $\pi$ after taking action $a$ in state $s$. Following policy $\pi$, $Q^{\pi}$ can be computed by the Bellman equation:
\begin{equation}\label{Qfunc}
\begin{split}
\scalemath{0.87}{
Q^{\pi}(s_t, a_t)  = \mathbb{E}_{s_{t+1}\sim p}[r(s_t, a_t)+ \gamma\mathbb{E}_{a_{t+1}\in{A}}[Q^{\pi}(s_{t+1}, a_{t+1})]],
}
\end{split}
\end{equation}
where $A$ represents the action space and $p$ is the state distribution. Let $Q^{*}(s, a)$ be the optimal action-value function. RL algorithms aim to find an optimal policy $\pi^*$ such that $Q^{\pi^*}(s, a)=Q^{*}(s, a)$ for all states and actions. The learned policy should predict the action at every step, guiding the manipulator to reach the goal while avoiding collisions. 
The detailed formulation is described below.
\begin{itemize}
    \item \textbf{States}: A feature vector is used to describe the continuous state, including the robot's proprioception, the obstacle and goal information in the environment. We restrict the orientation of the gripper as orthogonal and downward to the table, so three joints out of six in our UR5e platform are active in the learning process. At each step $t$ we record the $i$-th joint angles $j_i$ for $i=1,2,3$ in radians and the end-effector's position $(x^{ee}, y^{ee}, z^{ee}) \in \mathbb{R}^3$ as the proprioception: ${\rm proprio}=(j_1, j_2, j_3, x^{ee}, y^{ee}, z^{ee}) \in \mathbb{R}^6$. Then, we estimate the obstacle's position in task space and use a bounding box to describe it: ${\rm obs} = (x^{o}_{min}, x^{o}_{max}, y^{o}_{min}, y^{o}_{max}, z^{o}_{min}, z^{o}_{max}) \in \mathbb{R}^6$. The goal is described as a point in the task space: ${\rm goal}=(x^g, y^g, z^g) \in \mathbb{R}^3$. Finally, the state feature vector can be represented as: $s=({\rm proprio}, {\rm obs}, {\rm goal}) \in \mathbb{R}^{15}$. 
     \item \textbf{Actions}: Each action is denoted by a vector $a \in ([-1, 1])^3$, which represents the relative position change for the first three joints. The corresponding three joint angle changes are $0.125 a$ rads.
    \item \textbf{Rewards}:
    A success is reached if the Euclidean distance between the end-effector and the goal ${\rm dis}({\rm ee}, g)<{\rm err}$, where ${\rm err}$ controls the reach accuracy. Given the current state and the taken action, if the next state is not collision-free, then a severe punishment is given by a negative reward $r=-10$. If the next state results in a success, we encourage such a behavior by setting the reward $r=1$. In other cases, $r=-{\rm dis}({\rm ee}, g)$ to penalize a long traveling distance. An episode is terminated when the predefined maximum steps or a success is reached. 
\end{itemize}
 Since the goal and obstacles are static in each episode, the state transition function $f_s$ is defined by $\mathcal{FK}$. Given the state $s_{i}$ and action $a_{i}$, the next state~$s_{i+1}$ can be calculated by the function~$f_s$ under the position controller: $s_{i+1} = f_s(s_i, a_i)$. We make a natural assumption that the goal states are reachable. 
 
 Since the transition function is known, our off-policy reinforcement learning framework can also be regarded as model-based. However, unlike traditional model-based RL directly involving the model in policy optimization and decision-making, we mainly use the model to generate demonstrations to facilitate our “model-free” RL agent learning better from its own experience. Besides, most of the model-based RL heavily depends on the model accuracy to avoid suboptimal performance, our method focuses more on the model-free exploration process. Thus, we have a lighter dependence on the model accuracy. The model we used is mainly for demonstration generation, and the self-imitation learning module helps the robot learn only from good experiences. 

\subsection{Experience-based Planning with PRM}\label{32}
In motion planning, in order to respond to new requests efficiently, past solutions are stored in memory and can be retrieved and repaired for later usage to speed up the planning process. This strategy is called experience-based planning  \cite{8794317, 9366973, 7139284}, where the experiences represent the previous solutions. However, in the context of RL, experiences refer to the past discrete decisions in the Markov Decision Process. It describes how the policy acts on a specific state and what the next state is under this action. Although most of these experiences are not part of the solutions, especially in the beginning exploration process, they still provide information to help understand the environment and the task. In this study, we generate demonstrations for SIL by planning on these experiences from the current policy.    

Many graph-search-based planning approaches can be used in our SILP+ to plan paths; here, we propose a PRM-based path planning as the planner. PRM is a multi-query SBMP algorithm, which exploits the fact that in SBMP, it is cheap to check whether a single robot configuration is in free space or not. The roadmap contains nodes and edges, in which a node represents a specific location, and an edge corresponds to a path connecting two nodes. After the roadmap has been generated, planning queries can be answered by connecting the user-defined initial and goal configurations.

The basic PRM is designed for static path planning applications, and it can realize multi-query path planning by constructing a global roadmap. In our environment, the location of the obstacle is randomly selected within a defined workspace in each episode, making the basic PRM inefficient, as the roadmap needs to be constructed in every episode. Hence, we build a directed graph on the set of visited collision-free states denoted by $\mathcal{S}_f$ in each episode. Each node in $\mathcal{S}_f$ corresponds to a state in MDP. These states are already collision-checked as they are selected from the RL experience; therefore, the nodes collision-checking process can be omitted and the efficiency of the algorithm has thus been improved. We denote the edge between nodes $s_i$ and $s_j$ as $e_{s_i\rightarrow s_j}$. Edges which have lengths greater than $d$ or intersect with obstacles are ignored. Finally, the graph $\mathcal{G}$ is formulated as follows:
\begin{equation}\label{roadmap}
\begin{split}
    \mathcal{G} &\triangleq (\mathcal{V}, \mathcal{E}),\\
    \mathcal{E} &=\left\{
    \begin{aligned}
    & \emptyset & & \text{if } d(s_i, s_j)>d \ \text{or}\ e_{s_i\rightarrow s_j} \in \mathcal{C}_{o} , \\
    & \{e_{s_i\rightarrow s_j}\mid s_i, s_j\in \mathcal{S}_f\} & & \text{otherwise},
    \end{aligned}
    \right.
\end{split}
\end{equation}
where $\mathcal{V} \in \mathcal{S}_f$ and $\mathcal{E}$ represent the nodes and edges in the graph, respectively, and $\mathcal{C}_{o}$ represents the edge collision space. 
We use \textit{A-star} as the local planner to extract paths in each episode, in which the end effector's positions in states are used to calculate the heuristic function. The adapted PRM-based path planning algorithm is illustrated in Fig. \ref{fig:PRM_path_planning} and explained as follows:
\begin{figure}[!hb]
\centering
	\includegraphics[scale=0.35]{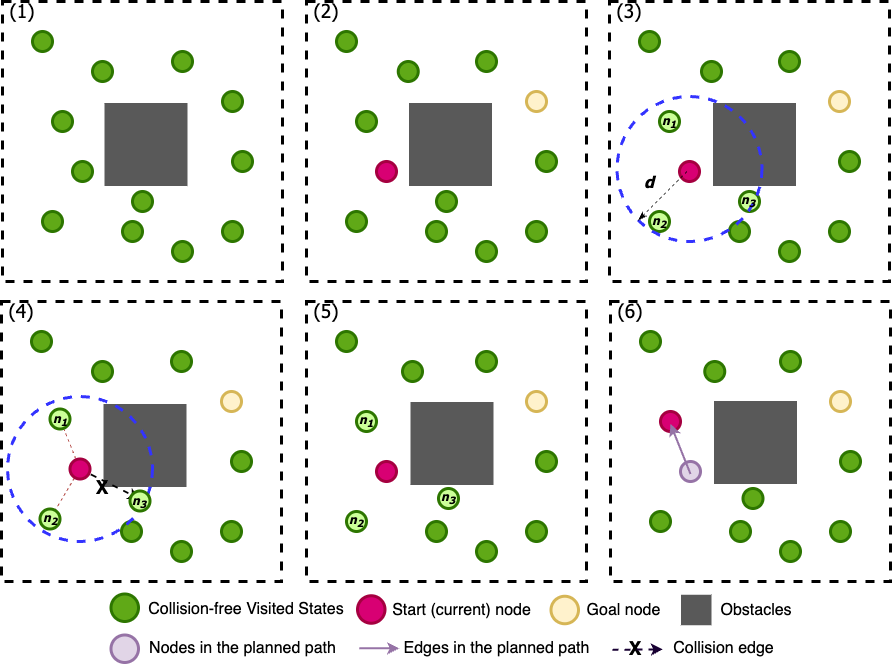}
	\caption{PRM-based path planning: (1) collision-free nodes construction. (2) start and goal configurations sampling. (3) candidate neighbors selection. (4) collision-checking on edges. (5) the local planner with A-star. (6) select the next node.}
	\label{fig:PRM_path_planning}
	\vspace{-2mm}
\end{figure}
\begin{enumerate}
    \item Collision-free nodes construction: Instead of randomly generating configurations in the workspace and selecting collision-free ones as the candidate nodes, we directly select the collision-free states from the experiences as candidate nodes and each node corresponds to a state in the MDP. The selected nodes are shown in Fig. \ref{fig:PRM_path_planning}(1). This step is also explained in Alg. \ref{CollisionFreeNodesconstruction}.
    
	\item  Start and goal configurations sampling: In the RL training process, the real goal probably cannot be reached, especially at the beginning of the training. Therefore, we randomly select $N$ start and goal pairs from the filtered collision-free nodes $\mathcal{S}_f$. Here, for the purpose of illustration, we use $N=1$ and illustrate the start and goal nodes in Fig. \ref{fig:PRM_path_planning}(2). Then, we set the start node as the current node $s_i$ and append it to the candidate path nodes $\mathcal{V}_e$. 
	\item  Candidate neighbors selection: To determine the candidate neighbors of node $s_i$, we use the Euclidean distance as the metric to choose all the nodes not in $\mathcal{V}_e$ of which the distances to $s_i$ are less than $d$, such as the nodes $n_1, n_2, n_3$ in Fig. \ref{fig:PRM_path_planning}(3). The selection of the distance $d$ is explained in \ref{appendixA}. We put them into a neighbors set. If no neighbors exist, we stop the planning process and continue with RL learning. 
	
	\item Collision-checking on edges: We check the edges between $s_i$ and the candidate neighbors using our collision checking module, which is explained in \ref{appendixB}. Candidate neighbors with collision edges are removed. As shown in Fig. \ref{fig:PRM_path_planning}(4), the edge between $s_i$ and $n_3$ intersects with the obstacle and thus is removed from the neighbor set. The method for edge collision-checking is based on subdivision, in which the intermediate linear interpolations of configurations are sampled based on $step\_size$. The edge between two nodes is collision-free if all of the intermediate configurations are checked collision-free.

	\item  The local planner with \textit{A-star}: We compute the cost of each candidate neighbor in the task space using heuristic functions. The cost of a candidate neighbor $nn$ is defined as: $f(nn) = h(nn) + g(nn)$, where functions $h$ and $g$ are the cost from $s_i$ to $nn$ and that from $nn$ to the goal, respectively. Both functions use Euclidean distance as the cost value.
	
	\item  Select the next node: The node with the least cost will be selected as the next start node, as shown in Fig. \ref{fig:PRM_path_planning}(6) and be appended to the candidate path nodes $\mathcal{V}_e$. The edge that connects the current node and the next node will be further processed when converting the path node $\mathcal{V}_e$ into demonstrations. Starting from the current node and repeating steps (3) to (6), the algorithm terminates when the final goal has been reached, the maximal running time has been used or no available neighbors can be found.
\end{enumerate}

\begin{algorithm2e}[t]

	\textbf{Input}: on-training policy $\pi$, environment $env$; empty Collision-free nodes $\mathcal{S}_f = \emptyset$\;
	\textbf{Output}: Collision-free nodes in $\mathcal{S}_f$, experience $\{S, A, S', R, Done\}$\;
	$S, A, S', R, Done = [\ ]$\;
	$s = env.reset$ \;
	\While {$episode\ not\ end$}{
		$a = \pi (s)$ \\
		$s', r, done, collision = env.step(a) $\\
		\While{collision}{
		$a = random \ action$ \\
		$s', r, done, collision = env.step(a) $\\
		}
	$S.append(s)$; $A.append(a)$; $S'.append(s')$ \\
			$R.append(r)$; $Done.append(done)$ \\
			$\mathcal{S}_f \leftarrow \mathcal{S}_f \cup \{ s' \} $ \\
	
		$s = s'$\\
    }	
	\caption{Collision-free nodes construction}
	\label{CollisionFreeNodesconstruction}
\end{algorithm2e}

\subsection{Online Generation of Demonstrations}
In order to learn from demonstrations with RL, we need to convert the path node $V_e=\{s_0, ..., s_n\}$ to MDP format for further imitation learning. Before proceeding, we make the following assumptions: (1) the forward kinematics function $f_s$ is given and based on $f_s$, the end effector's position is predictable using the robot's joint values. (2) the inverse model $f_a$ is given so that the action $a_i = f_a(s_i, s_{i+1})$ that controls the agent from one state to another state is accessible. Then, we illustrate the pseudo-code for online demonstrations generation in Alg.~\ref{PRM-algorithm}, in which the objective is to convert the state-based nodes $V_e$ to $(s, a, s', r, done)$ tuples and save them in a demonstration replay buffer $D_{demo}$. The conversion iterates the successive nodes $s$ and $s'$ and calculates the action $a$ using the inverse model $f_a$. Based on $(s, a, s')$, the reward $r$ and the Boolean value $done$ can be calculated. However, the predicted action may be out of the action space $A$. In this case, we insert an extra node between the states with a half value of $a$, and the new next state is calculated by $f_s$. This process is repeated until the action is within action space $A$. We save the constructed demonstrations in $D_{demo}$ for policy learning. 
\setlength{\textfloatsep}{0pt}
\begin{algorithm2e}[!ht]
	\textbf{Input}: Planned nodes $V_e = \{s_0, s_1, ..., s_n\}$ in episode~$e$; environment $env$; action space $A$; action model function $f_a$ and forward kinematics function $f_s$\;
	\textbf{Output}: Updated demonstration replay buffer $D_{demo}$\;
	\textbf{Demonstrations Generation}:\\
	\For{$i$ in \textsc{length}($V_e$)}
	{
		$s = V_e[i]$ \;
		$s' = V_e[i+1]$ \;
		$a = f_a(s, s')$ \;
		\While{$a$ $\not\in$ $A$}{
			$(s'', a') = \textsc{InsertNodes}(s, s', a)$\;
			$r' = env.\textsc{reward}(s, s'')$ \;
			$D_{demo}.\textsc{push}((s, a', s'', r'))$\;
			$ s = s''$ \;
			$a = f_a(s'', s')$ \;
		} 
		$r = env.\textsc{reward} (s, s')$ \;
		$D_{demo}.\textsc{push}((s, a, s', r))$
	}
	\SetKwFunction{FMain}{InsertNodes}
	\SetKwProg{Fn}{Function}{:}{}
	\Fn{\FMain{$s$, $s'$, $a$}}{
		$a = a/2$\;
		$s' = f_s(s, a)$\;
		\While {$a$ $\not\in$ $A$}{
			$(s', a) = \textsc{InsertNodes}(s, s', a)$\\
		}
		\KwRet $s'$, $a$\;
	}
	\caption{Demonstrations Generation in SILP+}
	\label{PRM-algorithm}
	\vspace{5mm}
\end{algorithm2e}

\begin{figure}
\centering
	\includegraphics[scale=0.8]{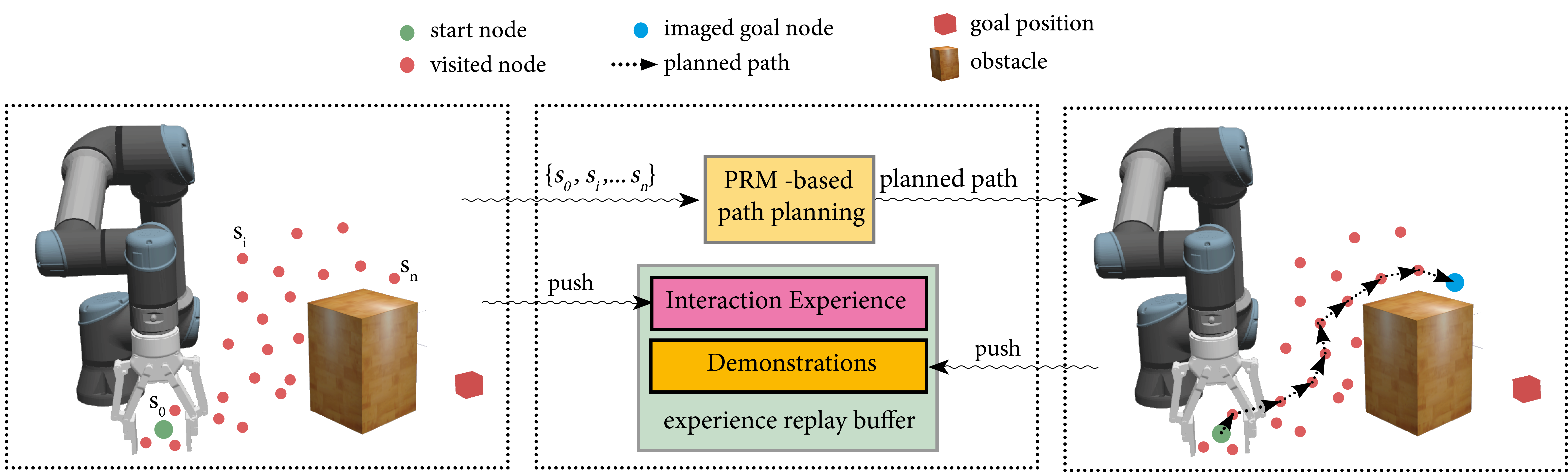}
	\caption{Self-imitation learning by planning plus}
	\label{fig:PRM-architecture}
	\vspace{2mm}
\end{figure}

The overall structure of SILP+ is depicted in Fig.~\ref{fig:PRM-architecture}. At the end of each training episode, we obtain the visited states, as shown in the left sub-figure. These experiences are stored in the interaction experience replay buffer. Based on the visited states, we plan paths using PRM-based planner, and the result is shown as the directed-dotted lines in the right sub-figure. Then, these paths are converted as demonstration tuples saved in the demonstration replay buffer for further imitation learning. 
\subsection{Self-Imitation Learning}
Self-imitation learning is an RL method that encourages actions whose returns were higher than the expectation \cite{oh2018self}. It was proven to be able to improve the performance of actor-critic methods in several discrete control tasks. One of the challenges in SIL is the difficulty to perform the task if the exploration performs poorly. For instance, a random exploration never generates a good experience within a reasonable time. Therefore, the policy cannot benefit from imitating the good experience when there is no good experience. To this end, we propose to combine the SIL with experience-based planning, called SILP+, in the context of motion planning. The planning module provides the demonstrations based on the visited experience, while SIL pushes the policy update toward the demonstrations. 

The method of embedding demonstration into SIL is adapted from~\cite{8463162}, where demonstrations are stored in a separate replay buffer $D_{demo}$, along with an interaction experience replay buffer $D_{\pi}$ that contains the interaction experiences. At each training step, we update the policy with $N_D$ and $N_{\pi}$ examples sampled from $D_{demo}$ and $D_{\pi}$, respectively. The guidance from demonstrations is implemented with a behavior cloning loss as shown below:
\begin{equation}\label{loss bc}
L_{bc} = \sum\nolimits_{i=1}^{N_{D}} \,\norm{\pi(s_{i}|\theta^{\pi})-a_{i}}^{2},
\end{equation}
where $a_i$ and $s_i$ are the action and state sampled from buffer $D_{demo}$. The $\theta^{\pi}$ represents the learning parameters in the policy. The policy imitates the good choices from the demonstrations by adding the behavior cloning loss to the objective $J$, which is weighted with hyperparameters $\lambda_1$ and $\lambda_2$, as shown below: 
\begin{equation}\label{objective}
\lambda_1\nabla_{\theta_{\pi}}J-\lambda_2 \nabla_{\theta_\pi}L_{bc},
\end{equation}

In order to avoid learning from imperfect demonstrations, a $Q_{filter}$ is employed in SILP~\cite{9561411} to prevent adding behavior cloning loss when the policy's action is better than the action from the demonstrations. However, experience replay buffer-based off-policy RL methods, such as DDPG and SAC, prone to the extrapolation error in Q function approximation. They struggle to learn when the data is different from the current policy's data distribution. For example, when updating the target Q values in DDPG and SAC, the next state $s'$ and action $a'$ from the target policy are involved. It risks to obtain unreliable Q value estimations as these state-action pairs are likely to be unfamiliar to the policy and not existed in the replay buffer. Fujimoto et al.~\cite{fujimoto2019off} demonstrated that the actor-critic algorithms deteriorate when the data is uncorrelated, and the value estimation produced by the Q-network diverges. With these considerations, Q filter used in \cite{8463162} could introduce the extrapolation error and lead to an incorrect filter for the action gap-guided RL training. 

Since we use planning as the source of demonstrations, it is natural to utilize models to form a more reliable Q filter. To this end, we replace the Q filter with a predicted reward filter, in which instead of comparing the state-action values between the policy and demonstrations, we predict the rewards $f_r$ based on the actions from the policy and demonstrations. The objective in (\ref{objective}) is changed to:
\begin{equation}\label{new obj}
\lambda_1\nabla_{\theta_{\pi}}J-\lambda_2 R_{filter} \nabla_{\theta_\pi}L_{bc},
\end{equation}
where
\begin{equation}\label{R filter}
R_{filter} =\left\{
\begin{aligned}
& 1 & & \text{if}\quad R(s_i, a_i)>R(s_i, \pi(s_i)),\\
& 0 & & \text{otherwise}.
\end{aligned}
\right.
\end{equation}
where
\begin{equation}\label{r demon}
R(s_i, a_i) = r_i+\gamma f_r(s_{i+1}, \pi(s_{i+1})) + \gamma^2 f_r(s_{i+2}, \pi(s_{i+2})) + ... + \gamma^k f_r(s_{i+k}, \pi(s_{i+k})), 
\end{equation}
and $(s_i, a_i, s_{i+1}, r_i)$ is one of the MDP tuple batches in $D_{demo}$. For computing $R(s_i, \pi(s_i))$ in (\ref{R filter}), we use (\ref{r demon}) with $r_i$ being replaced with $f_r(s_i, \pi(s_i))$. Correspondingly, next state $s_{i+k}$ should be predicted based on $f_s$.
 While larger values of $k$ can potentially yield better performance \cite{heess2015learning} by capturing longer-term dependencies, they often require more time during training due to the increased number of time steps to be considered. Given the complexity of our problem domain and the objectives of our study, we opted for the one-step return $k=1$ to strike a balance between performance and training efficiency. However, it is straightforward to use $k>1$.

\begin{figure}[!h]
\centering
	\includegraphics[scale=0.15]{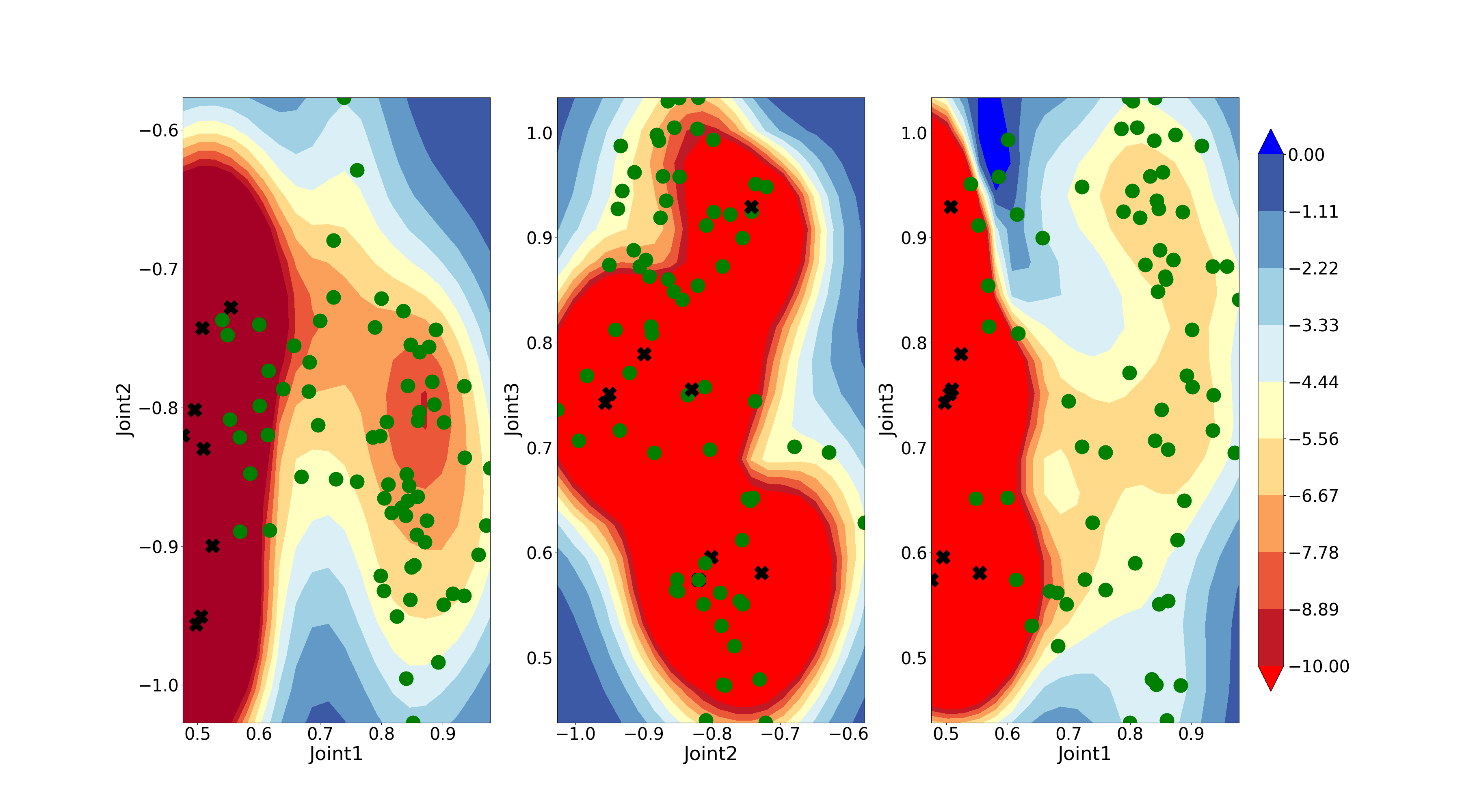}
	\caption{An example of modeling the reward function with a Gaussian process in three views on the 3D joint space: The green points represent visited collision-free nodes and the black, crossed points represent the collision nodes. Different color levels represent the modeled reward values which increase from red (penalty) to blue (high reward). The units of the axes coordinates are radians. The modeled reward distribution can guide the RL policy to explore regions with higher rewards. Please note that the positions of the points are based on a 2D projection from the 3D joint space.}
	\label{fig:gp} 
	\vspace{5mm}
\end{figure}
\subsection{Gaussian Process-Guided Exploration}
In SILP \cite{9561411}, when a collision happens during training, the agent would go one step back to the before-collision state and randomly select another action to continue. Under this strategy, we observed a scenario that the policy is likely to get stuck on selecting random collision-free actions near the obstacles until running out of the steps. This issue is due to the limitation of random sampling, which need to take an impractical amount of samples to cover the whole space. Therefore, the agent is easy to run out of the pre-defined steps before it encounters a valid action. To tackle this problem, we propose Gaussian-process-guided exploration to improve the exploration quality near the collision regions. The main idea is to model the reward landscape based on the collected experiences when the collision happens and select the action with the most promising reward. 

We use the Gaussian process regression to approximate the function $f_m$ which maps from the experienced states to rewards. First, we assume $f_m$ is distributed as a Gaussian process: $f_m \sim \mathcal{GP}(0, k(x, x'))$, with zero prior mean and the covariance function $k: \mathcal{X} \times \mathcal{X} \to \mathbb{R}$. The output $y(x)$ of the function $f_m$ at input $x$ can be written as $y(x)=f_m(x)+\epsilon$, with the Gaussian noise $\epsilon \sim \mathcal{N}(\epsilon; 0, \sigma^2_{no})$. After having collected $n$ observations, denoted by $D_n = \{\textbf{\textit{x}}_n, \textbf{\textit{y}}_n\}=\{x_1, ...,x_n, y_1, ..., y_n\}$, the predictive distribution at location $x$ is given by
\[
    p(f_m \mid \mathcal{D}_n, x) = \mathcal{N}(f_m(x); \mu(x \mid \mathcal{D}_n), \sigma^2(x \mid \mathcal{D}_n)),
\]
with the predictive mean $\mu(x \mid \mathcal{D}_n)=\textbf{\textit{k}}^T_n(x)[K_n+\sigma^2_{no}I]^{-1}\textbf{\textit{y}}_n$ and the predictive variance $\sigma^2(x \mid \mathcal{D})=k(x,x)-\textbf{\textit{k}}^T_n(x)[K_n+\sigma^2_{no}I]^{-1}\textbf{\textit{k}}_n(x)$~\cite{10.5555/1162254}, where the entries of the vector $\textbf{\textit{k}}_n(x) \in \mathbb{R}^n$ are $[\textbf{\textit{k}}_n(x)]_i=k(x_i, x)$, the entries of the Gram matrix $K_n \in \mathbb{R}^{n \times n}$ are $[K_n]_{i, j}=k(x_i, x_j)$, and the entries of the vector of observations $\textbf{\textit{y}}_n\in \mathbb{R}^n$ are $[\textbf{\textit{y}}_n]_i=y_i$.
When there is a collision during training, we collect all of the states and rewards in the episode and fit the states and rewards as the input and output in $f_m$. Then we use the Monte Carlo planning to select the intended action: first, $n$ random state configurations are sampled; then, the rewards are predicted for these configurations using the learned model $f_m$; later, we scale the reward into a sampling probability within $[0, 1]$ and make sure that the sum of those probabilities is 1; finally, we choose one of those configurations under the scaled probabilities as our desired next state and retrieve the corresponding action. The kernel in Gaussian process regression we used is: Matern 5/2 \cite{rasmussen2003gaussian}. We depicted an example of fitted reward distributions in Fig. \ref{fig:gp}. From the figure, we can see that the experienced collision regions have lower rewards expectation and there are small probabilities to choose the next action in the vicinity of these regions.

\section{Experiments}\label{Experiments}
We conducted experiments in both simulations and real robot settings to answer the following questions: \textbf{(1)} Under the same environment and task settings, does SILP+ perform better than other baseline algorithms in terms of success rate and sample efficiency? \textbf{(2)} Will SILP+, which additionally includes an online demonstration generation step increases computation burden and leads to a slower training process? \textbf{(3)} How do collisions affect the learning performance? \textbf{(4)} How do the Gaussian-guided exploration and extrapolation errors affect the performance? \textbf{(5)} Can the policy learned in simulation transfers well to a physical robot where noise and uncertainty exist? 
\subsection{Task and Training Setup}
\begin{figure}[!t]
	\centering
	\begin{subfigure}[t]{0.45\textwidth}
		\centering
		\includegraphics[width=\textwidth]{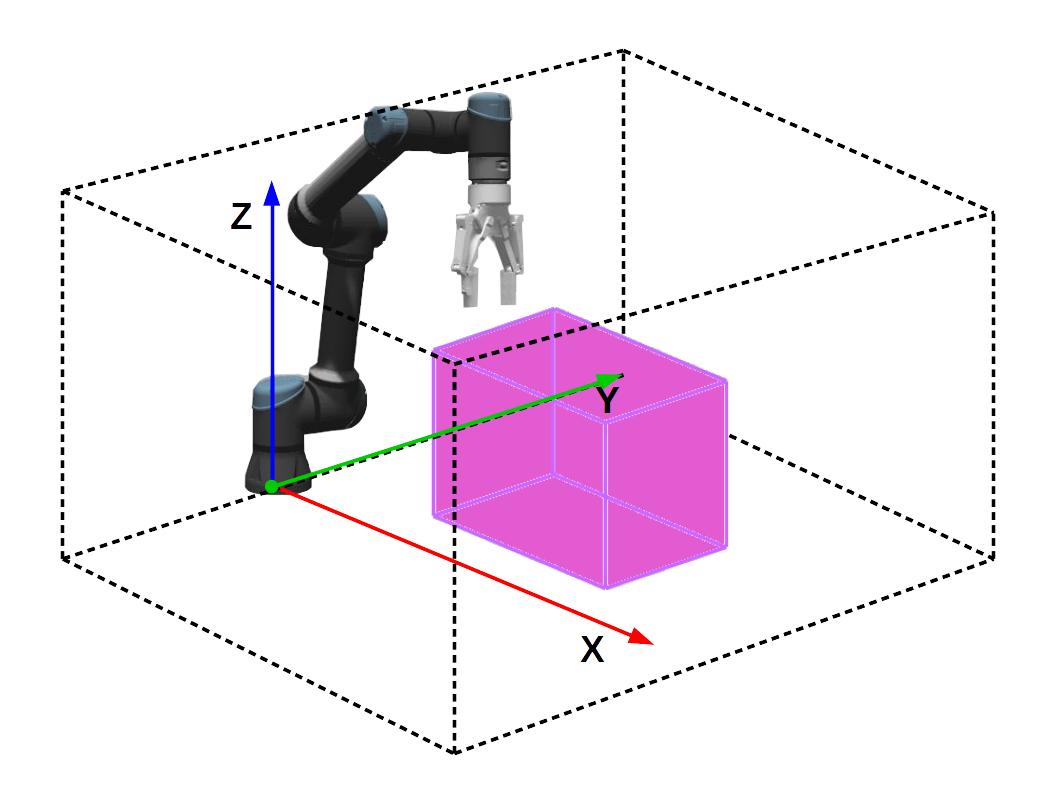}
		\caption[Task workspace]%
		{{\small Task workspace}}    
		\label{fig:task-workspace}
	\end{subfigure}
	\quad
	\hspace{.01in}
	\begin{subfigure}[t]{0.48\textwidth}  
		\centering 
		\includegraphics[width=\textwidth]{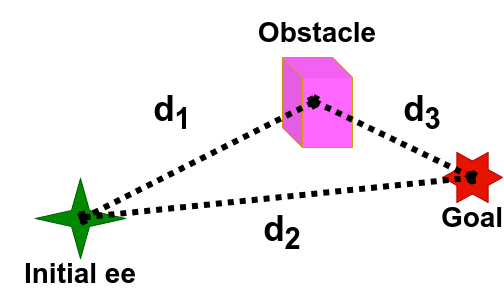}
		\caption[Position relationship in initialization]%
		{{\small Position relationship in initialization}}    
		\label{fig:task-relation}
	\end{subfigure}
	\caption[ws illustration]
	{\small (a) Task workspace: the transparent region bounded with dashed lines is the reachable workspace of the end effector and the goal; the purple region represents the region for the obstacle; (b) Position relationship in initialization: \mbox{$d_2>d_1>d_3$}.} 
	\label{fig:ws}
	\vspace{4mm}
\end{figure}

We used Gazebo with an ODE physics engine as the simulator for training policies, in which a 6 DoF robot arm UR5e is equipped with a Robotiq-2f-140 gripper to accomplish long horizon motion planning tasks. The workspace for the end-effector is restricted to $x\in[0, 0.8]$m, $y\in[-0.3, 0.8]$m, $z\in[0, 0.6]$m to simplify the tasks and avoid unnecessary collisions. A box with the width and height of 0.2m and 0.3m respectively is used as an obstacle in the task. The obstacle's position (the center of the mass) is limited to $x\in[0.3, 0.7]$m, $y\in[0.1, 0.4]$m, $z=0.15$m (see the purple region in Fig.~\ref{fig:task-workspace}). The initial arm pose and the goal pose were restricted within the reachable workspace of the end-effector, as mentioned before. In order to balance the number of collision and non-collision interactions, we restrict the initial pose, goal pose and obstacle position to satisfy $d_2>d_1>d_3$, where $d_1$ is the Euclidean distance between the initial end-effector's position and the obstacle's position, $d_2$ is the Euclidean distance between the initial end-effector's position and the goal position, and $d_3$ is the Euclidean distance between the obstacle's position and the goal position. This means that the robot needs to learn a generalized policy to reach the target from different directions while avoiding the obstacle. The relative positions are projected in 2D and depicted in Fig.~\ref{fig:task-relation}. 

\subsection{Baseline Comparison}
We employed two state-of-the-art off-policy RL algorithms as our basic baselines: deep deterministic policy gradient (DDPG) \cite{lillicrap2015continuous} and soft actor-critic (SAC) \cite{haarnoja2018soft, haarnoja2018soft2} to evaluate SILP+. We call them \textit{DDPG-SILP}+ and \textit{SAC-SILP}+, respectively. The design of the neural networks and hyperparameters for DDPG and SAC, as well as the training details can be found in \cite{9561411}. 
The procedure of SILP+ is described in detail in Section \ref{32}, above.

We compared the success rate (the number of successful task attempts divided by the total number of attempts) and training time for the following methods:
\begin{itemize}
	\item \textbf{DDPG}: \small{DDPG combined with a dense reward.}
	\item \textbf{DDPG-HER}: \small{DDPG combined with hindsight experience replay (HER)~\cite{andrychowicz2017hindsight}. }
	\item \textbf{DDPG-Demon}: \small{DDPG with experience replay buffer that contains demonstrations from online planning.}
	\item \textbf{DDPG-SILP+}: \small{DDPG combined with SILP+.}
	\item \textbf{SAC}: \small{SAC combined with a dense reward.}
	\item \textbf{SAC-HER}: \small{SAC combined with HER.}
	\item \textbf{SAC-Demon}: \small{SAC with experience replay buffer that contains demonstrations from online planning.}
        \item \textbf{SAC-SILP}: \small{SAC combined with SILP.}
	\item \textbf{SAC-SILP+}: \small{SAC combined with SILP+.}
	\item \textbf{PRM-0.1}: \small{PRM with planning time limited to 0.1 second. }
	\item \textbf{PRM-1}: \small{PRM with planning time limited to 1 second. }
	\item \textbf{BC-SAC-SILP+}: \small{Behavior cloning with demonstrations collected from SAC-SILP+ policy's rollout. }
	\item \textbf{BC-PRM}: \small{Behavior cloning with demonstrations collected from PRM in \textit{Moveit} under the planning time threshold of 1 second. }
\end{itemize}

For HER, the number of imagined goals is four for each visited state. The DDPG-Demon and SAC-Demon methods impose demonstrations into the regular experience replay buffer. Similar to SILP+, those demonstrations come from online PRM planning. The difference between SILP+ and Demon is that Demon does not use a behavior cloning loss-based SIL to specifically learn from good experience. Instead, Demon is more of a data argumentation method. PRM-0.1 and PRM-1 were implemented in Moveit under the planning time limitation of 0.1 second and 1 second, respectively. We used 10$k$ demonstrations to train the BC model. The demonstrations in BC-SAC-SILP+ were collected by the well-trained SAC-SILP+ policies, and the demonstrations in BC-PRM were collected by PRM through Moveit with planning time limited to 1 second. 

For DDPG- and SAC-based methods, we set the training epoch to 1K. Here, the epoch represents certain iterations of updates in parameters. Each epoch contains ten episodes. The results are summarized in Table \ref{tab:bl}. The training time means the wall time for the defined 1K epochs, which is summarized from three trained policies with different seeds.
The final success rates are measured under the three trained policies; each policy being tested 1K times. The planning time is the accumulated time for rolling out the learned policy, which is also summarized from the three trained policies, and each policy has been rolled out 1k times. Note that the PRM methods do not have training time as online planners do. The training time for BC-based methods comprises the data collection and policy training time.
\begin{table}[!ht]
\renewcommand\arraystretch{1.2}
\small
	\caption{Success rate, training, and planning time of SILP+ compared with other methods. Results are collected in simulation and averaged: Each method was randomly initialized three times, trained and then tested
    on 1000 reaching trials.}
	\label{tab:bl}       
\begin{tabular}{llll}

	\specialrule{0.1em}{0pt}{3pt}
	Algorithms    & Success Rate       &  Training Time (h:m:s)                    & Planning Time (s)    \\
	\specialrule{0.1em}{3pt}{3pt}
    DDPG          & 0.763 \space (0.046)   &  08.17.53 \space (00.23.26)   & 0.110 \space (0.017)\\
    DDPG-HER      & 0.532 \space (0.257)   &  11.15.01 \space (03.18.38)   & 0.126 \space (0.024)\\
    DDPG-Demon    & 0.707 \space (0.039)   &  08.07.51 \space (00.13.21)   & 0.107 \space (0.007)\\
    DDPG-SILP+     & \textbf{0.954} \space (0.021)   &  \textbf{07.35.11} \space (00.13.50) & 0.114 \space (0.009)  \\
    \hline
    SAC           & 0.864 \space (0.066)   &  07.15.24 \space (00.48.00)   & 0.116 \space (0.010)\\
    SAC-HER       & 0.902 \space (0.008)   &  07.25.42 \space (01.04.36)   & 0.113 \space (0.012)\\
    SAC-Demon     & 0.925 \space (0.021)   &  \textbf{05.19.24} \space (00.06.45)   & 0.121 \space (0.008)\\
    SAC-SILP     & 0.944 \space (0.004)   &  06.35.07 \space (00.09.25)   & 0.118 \space (0.003)\\
    SAC-SILP+      & \textbf{0.973} \space (\textbf{0.002})   &  05.22.58 \space (\textbf{00.04.26}) & 0.130 \space (0.008) \\
    \hline
    PRM-0.1      & 0.749 \space (0.038)   &  * & 0.134 \space (0.004) \\
    PRM-1      & 0.772 \space (0.067)   &  * & 1.028 \space (0.005) \\
    \hline
    BC-SAC-SILP+      & 0.967 \space (0.003)   &  00.38.03 \space (00.02.25) & 0.137 \space (0.008) \\
    BC-PRM      & 0.464 \space (0.015)   &  01.32.56 \space (00.01.59) & 0.144 \space (0.010) \\
	\specialrule{0.1em}{3pt}{0pt}
\end{tabular}
\vspace{4mm}
\end{table}

From the success rate column, we observed that SAC-SILP+ achieved the highest success rate (0.973) and the lowest standard variance (0.002) compared with other methods. The success rates in the DDPG spectrum are lower than SAC spectrum methods, but SILP+ can boost DDPG's performance to the same level of SACs, as we can see from the data in DDPG-SILP+ and SAC-SILP+. In addition, traditional PRM methods performed worse than our SILP+ methods, although the increased planning time could slightly improve the performance. Not surprisingly, BC methods depend heavily on the quality of the expert demonstrations. The demonstrations' quality in SAC-SILP+ is better than in PRM's. The reason is that the planned paths in PRM can be partly beyond the workspace or suboptimal because of singularities. As a result, BC with SAC-SILP+ demonstrations and PRM demonstrations achieved success rates of 0.967 and 0.464 respectively under the same test configurations. 

From the training time column, we found that SAC-Demon took the least training time, but SAC-SILP+ used a similar amount of training time while the standard deviation is the lowest. SAC-SILP+ and DDPG-SILP+ have achieved 28\% and 8\% less on training time compared to SAC and DDPG methods. Although the planning module intertwines with the learning process, there is no additional substantial computation load on the main program. This is due to the following facts: (1) the planning process is based on the visited states and it has been accelerated with the elimination of collision-checking on the candidate nodes; (2) the most computationally expensive step is the interaction with the environment and the planner involves a small proportion of the total computation.

In addition, HER-related methods are more unstable than others in our task as they have the highest standard deviation variance among the compared algorithms. For BC-related methods, we calculated the training time as a sum of both data collection and BC model training time. The training time for BC methods is much lower than RL-based methods. However, we should notice the assumption that the expert was given before BC training. The access to the expert also takes time and effort. In terms of the planning time, learning-based methods take similar amounts of time for rolling out the policies. The situation for PRM methods is different as we can define the desired planning time for the planner. The longer time we allocate, the higher success rate we can expect. We found that if we limit the planning time in PRM to 0.1 second, the same level of planning time in our learning-based methods, the mean success rate is 0.749, which is much lower than the result in DDPG-SILP+ or SAC-SILP+. From this perspective, SILP+ is superior to the traditional PRM method.

\begin{figure}
\centering
	\includegraphics[scale=0.3]{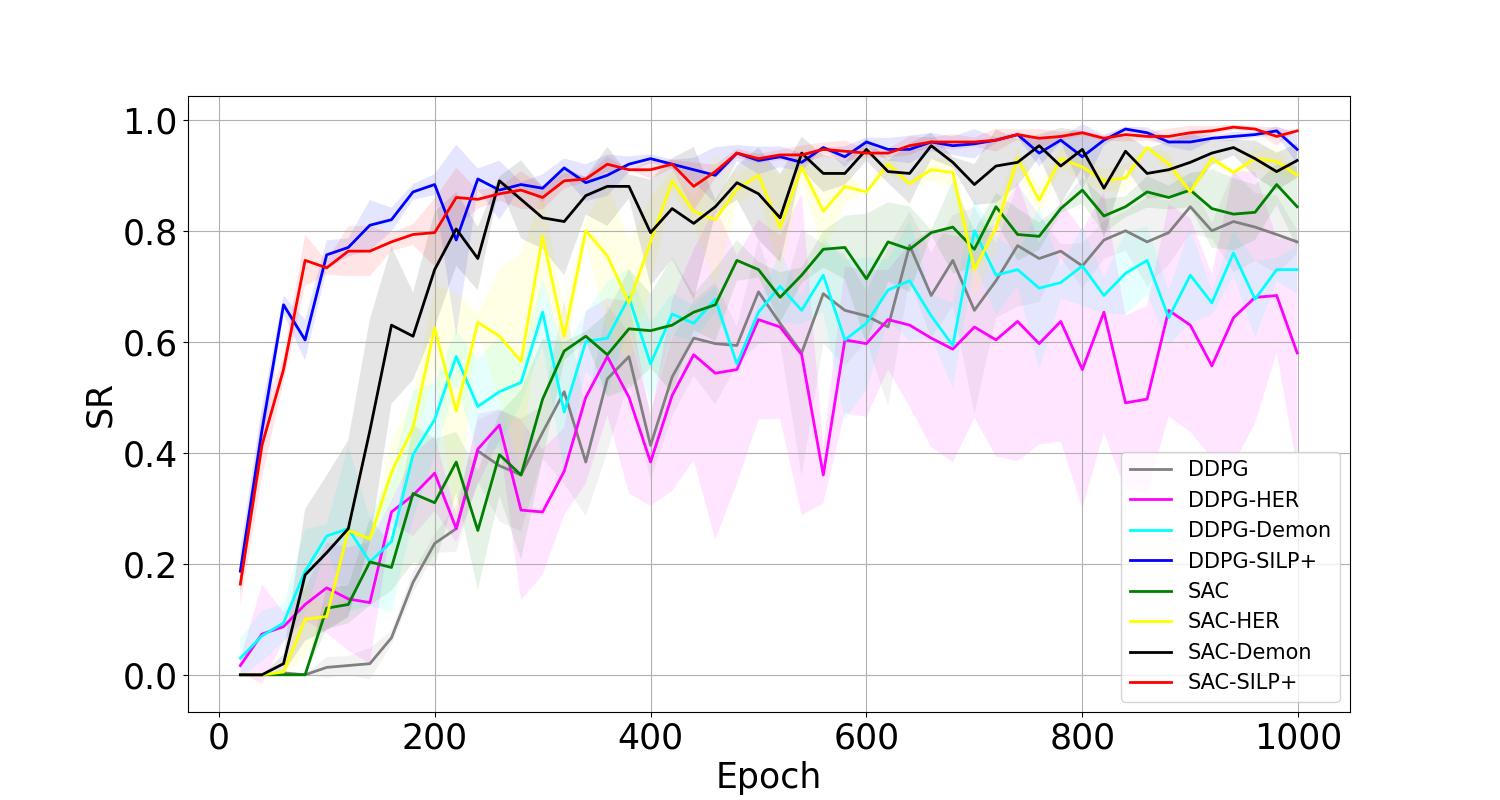}
	\caption{Success rates during training, the data is recorded every 20 epochs. The solid line represents the mean value and the transparent region represents the standard deviation range under three randomly chosen seeds.}
	\label{fig:sr} 
	\vspace{5mm}
\end{figure}

We also compared the performance during the training in terms of success rate, as shown in Fig. \ref{fig:sr}. The curves indicate that our SILP+ related methods DDPG-SILP+ and SAC-SILP+ perform better than other methods in the very beginning of learning. The improvements slower down after 100 epochs, but the advantage of both methods remains significant until the end of training. 

The above baseline comparisons verified higher success rates and better training efficiency of SILP+ than other algorithms, including SILP. The main contributions of SILP+ compared to SILP include the Gaussian process-guided exploration near obstacles, the reward-based filter in the self-imitation learning framework, and the learning strategy in response to collisions. While the difference in success rates may appear small between SAC-SILP+ (0.973) and SAC-SILP (0.944), it is essential to consider the context. The success rate ceiling is 1, each incremental improvement becomes increasingly significant as we approach the maximum performance. Additionally, when we consider the efficiency aspect. We notice that the average training time for SILP+ has an improvement of 18.2\% compared to SILP. This significant reduction in training time serves as a clear and easily understandable indicator of the superior performance of SILP+ over SILP. 

In the next subsections, we further investigate how each of these elements contributes to better performance and training efficiency in SILP+.

\subsection{Collision Types Comparison}
It is commonplace to encounter failures during the learning process in motion planning tasks. Herein we consider scenarios where failing is undesirable but not catastrophic, such as, avoiding touching, moving fragile objects or immobile obstacles. In such scenarios, failures still provide a valuable source of information, which have normally been handled by user-defined penalties in reward functions. Yet, designing effective reward functions to avoid failures is difficult and usually requires domain knowledge. Besides designing a suitable reward function, the way of dealing with collision states in the RL framework also plays a vital role in the learning performance. We consider the following three methods to deal with the collisions: 
\begin{itemize}
    \item \textbf{type-0} (early-reset on collisions): The algorithm terminates the episode when a collision happens \cite{choi2021reinforcement} \cite{zhao2021reinforcement} and gives the accident a punishment in the reward function \cite{zhao2021reinforcement} \cite{sangiovanni2018deep}.
    
    \item \textbf{type-1} (continue as if nothing happens):  The algorithm continues the episode with another random but collision-free action; collision experiences are skipped and will not be added into the experience replay buffer for training.
    
    \item \textbf{type-2} (learn from collisions and successes):  The algorithm continues the episode with another random but collision-free action and adds all of the collision experiences into experience replay buffer for policy training.  
\end{itemize}

\begin{figure}
\centering
	\includegraphics[scale=0.30]{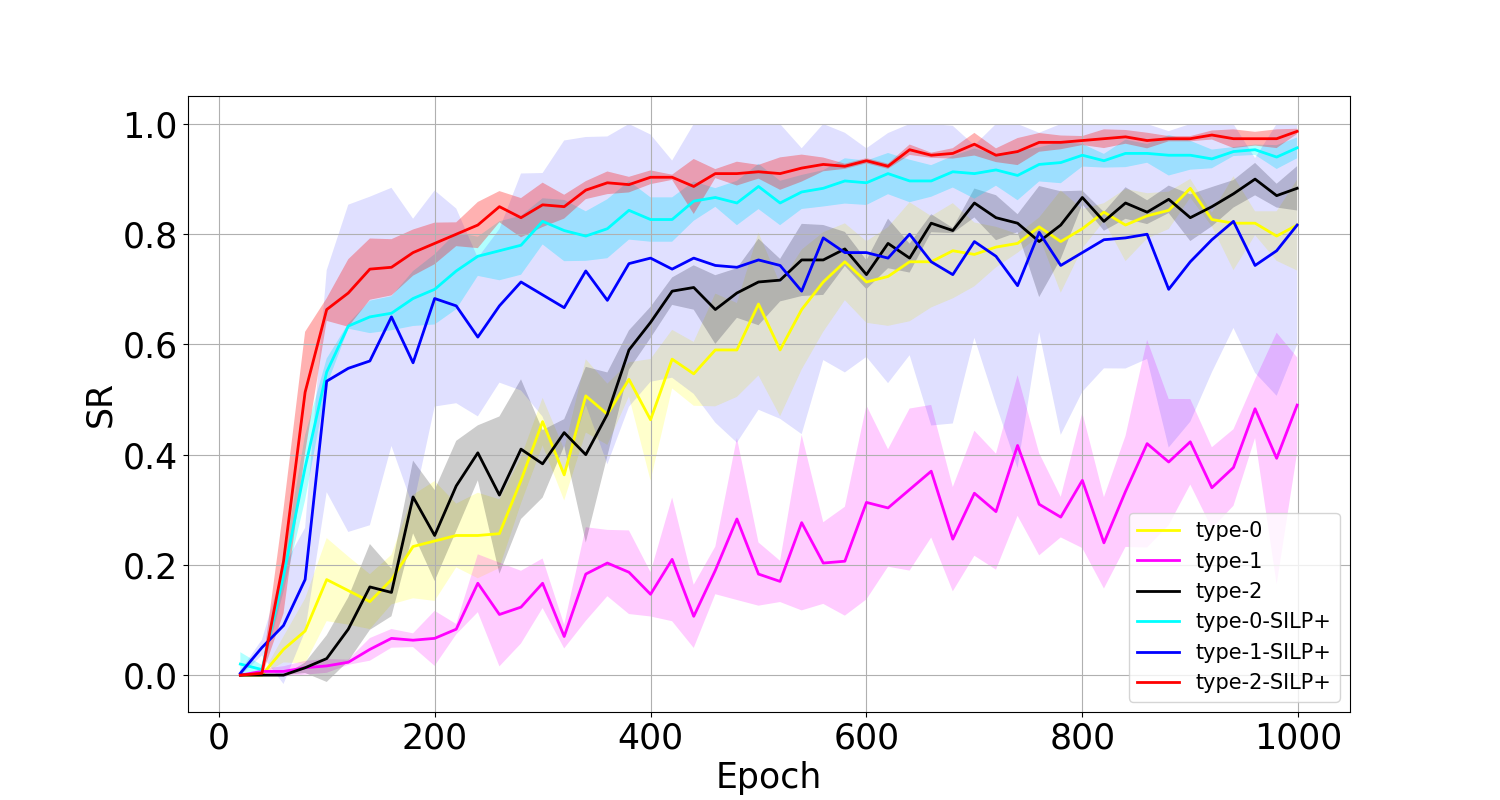}
	\caption{The success rate of different ways of dealing with collisions during training; the data is recorded every 20 epochs. The solid curve is the mean value, and the transparent region is the standard deviation under three randomly chosen seeds. Type-0, type-1 and type-2 are different types of methods dealing with collisions. If we use these collision methods with SILP+, we have type-0-SILP+, type-1-SILP+ and type-2-SILP+. }
	\label{fig:ce-type} 
	\vspace{5mm}
\end{figure}

We recorded the success rates during training with these three methods on pure SAC (type-0, type-1, type-2) and SAC-SILP+ (type-0-SILP+, type-1-SILP+, type-2-SILP+) and illustrated them in Fig. \ref{fig:ce-type}. The training time, the number of collisions, accumulated steps during training, and the success rates on trained policies (average with three seeds on 1K episodes) are summarized in Table \ref{tab:ce-type}.

The difference between type-0 and type-1 is that the latter will not terminate the episode when a collision happens. Instead, type-1 will randomly select collision-free actions to continue the episode. This results in a longer episode and training time, as shown in Table \ref{tab:ce-type}, the training time in type-1 has increased 70.3\% and 78.5\% in SAC and SAC-SILP+ compared to type-0. In addition, type-1 adds more experience and information near the obstacles, but the success rate of type-1 (0.423) in SAC is much lower than type-0 (0.819). The declined performance could attribute to the scarce random exploration near obstacles. In continuous space, these explorations are inadequate to help the policy gain a generalized understanding of the environment but confuse the policy. However, the situation is different when SILP+ gets involved; one can observe that all three types benefit from SILP+ in success rate and training time. Especially for the success rate of type-1, it is more than double that of SAC-SILP+, from 0.423 to 0.942. We interpret that the randomly explored states near the obstacle could enrich the nodes in PRM and help SILP+ plan better distributed demonstrations, thus alleviating the scarce experience problem near obstacles. 

The difference between type-1 and type-2 is that type-2 uses the collision experiences to update its policy while type-1 does not. From the success rate in Table \ref{tab:ce-type}, we see that type-2 performs better than type-1 in both SAC and SAC-SILP+ settings. In accordance, the training time decreases, especially in SAC methods. We conclude that the failure experience can boost the performance, which is also reflected in the comparison between type-0 and type-2.
\begin{table}[!t]
\renewcommand\arraystretch{1.2}
\small
	\caption{Success rates and training time under different collision methods (simulation)}
	\label{tab:ce-type}       
	\centering
\begin{tabular}{cccc}
	\specialrule{0.1em}{0pt}{3pt}
	\multicolumn{2}{c}{Methods} & Success Rate  &  Training Time (h.m.s) \\
	\specialrule{0.1em}{3pt}{3pt}

	\multirow{3}*{SAC}& type-0 & 0.819\space ($\pm$ 0.079) & 06.18.07\space ($\pm$ 00.40.44)\\
	\cline{2-4}
	~ & type-1 & 0.423\space ($\pm$ 0.113) & 10.43.53\space ($\pm$ 00.22.14)\\
	\cline{2-4}
	~ & type-2 & \textbf{0.865\space ($\pm$ 0.021)}& 07.12.06\space ($\pm$ 00.05.08) \\
    \hline
	\multirow{3}*{SAC-SILP+}& type-0 & 0.952\space ($\pm$ 0.027)  & 04.59.39\space ($\pm$ 00.06.08)\\
	\cline{2-4}
	~ & type-1 & 0.942\space ($\pm$ 0.015) & 06.45.16\space ($\pm$ 00.32.18)  \\
	\cline{2-4}
	~ & type-2 & \textbf{0.971\space ($\pm$ 0.010)}  & 06.10.02\space ($\pm$ 00.14.13)  \\
	\specialrule{0.1em}{3pt}{0pt}
\end{tabular}
\vspace{5mm}
\end{table}

From the analysis above, one observes that the positive feedback from the planned path in SILP+ also helps with learning. However, which information has more impact on the results? Negative feedback from failures or positive feedback from the planned demonstrations? The answer can be found when one compares the success rates between type-0 and type-2 (from 0.819 to 0.865) and type-0 and type-0-SILP+ (from 0.819 to 0.925). The improvements from negative and positive feedback are 0.046 and 0.133, respectively. Therefore, we interpret that the positive feedback from LfD is more important than the negative feedback in training an NMP.   

Here, we take the high success rate as our objective, so we selected type-2 to deal with collisions during the training process. However, if the efficiency has a higher priority, type-0 is also a good option as it can achieve comparable performance under the SILP+ algorithm but requires less training time.

\subsection{Gaussian-Process-Guided Exploration}
In this subsection, we did an ablation experiment to investigate the effect of the Gaussian-process-guided exploration. We compared SAC-SILP+, which was embeded with the Gaussian-process-guided exploration module, with another SAC-SILP+ that is without this module. We call them with-GP and without-GP, respectively. The success rate, training time and the number of collisions during training are summarized in Table \ref{tab:gp}. From the data, we can see that the final success rates are the same, but the number of collisions for with-GP has decreased to around 20\% of without-GP. Accordingly, the training time has been shortened by more than 1 hour. The greatly decreased collision number can be beneficial for safety-sensitive applications, especially in the robotics field.
\begin{table}[!ht]
\renewcommand\arraystretch{1.2}
\small
\centering
	\caption{Performance with Gaussian-Process-Guided Exploration (simulation)}
	\label{tab:gp}       
	\centering
\begin{tabular}{lll}
	\specialrule{0.1em}{0pt}{3pt}
	Performance           &  without-GP   &  with-GP  \\
	\specialrule{0.1em}{3pt}{3pt}
	SR                 &  0.973       & 0.973 \\
	Time (h.m.s)               &  6.28.50     & 5.22.58 \\
	Collision Number   &  30844        & \textbf{6153} \\
	\specialrule{0.1em}{3pt}{0pt}
\end{tabular}
\end{table}

\begin{figure}
\centering
	\includegraphics[scale=0.27]{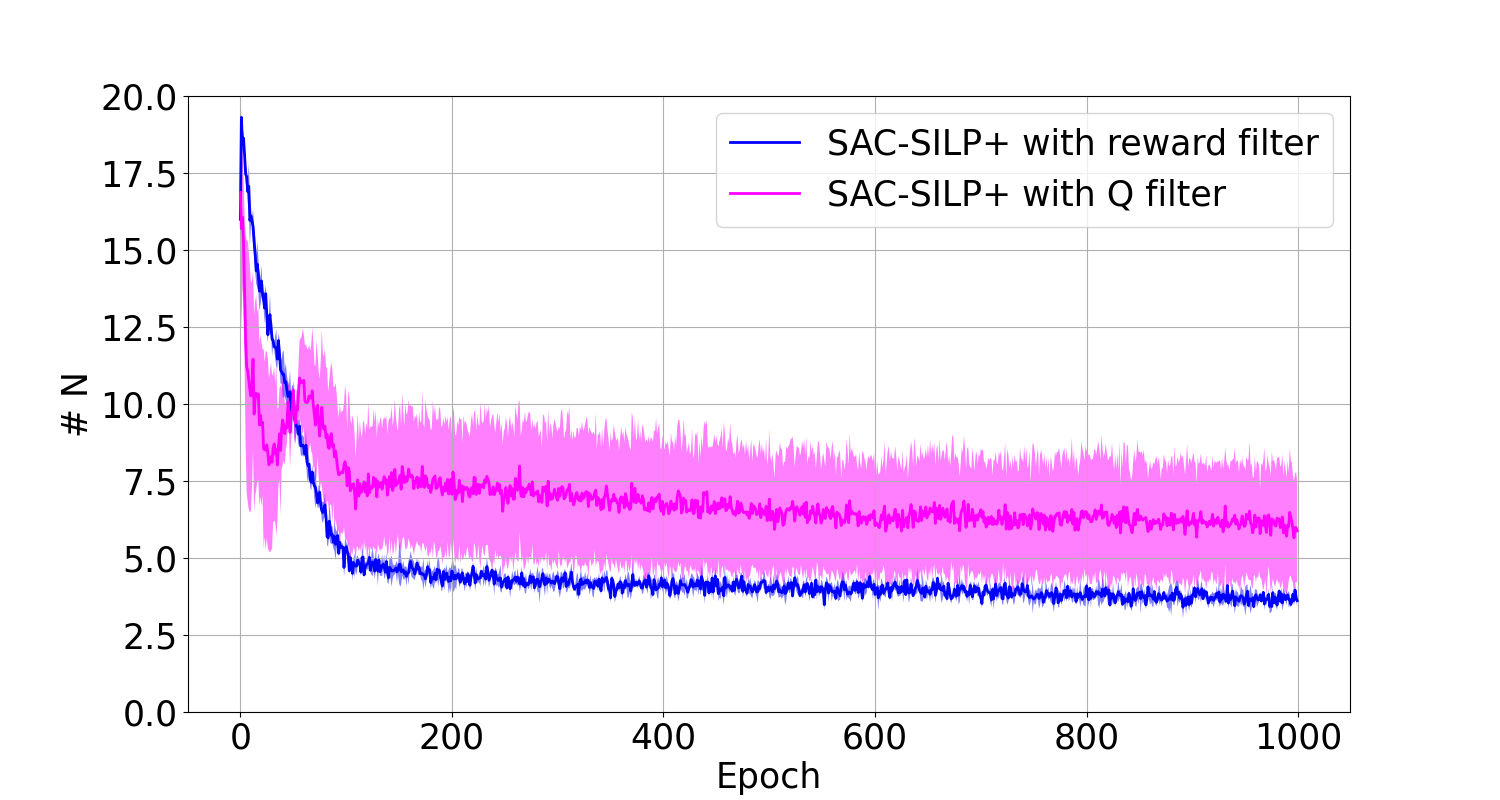}
	\caption{Number of actions (\#N) from demonstrations that perform better than the RL policy during training. In SAC-SILP+ (dark blue curve), the criterion for comparison is based on the rewards. In SAC-SILP+ with Q filter (ligh magenta curve) action-state values from the critic are used as the criterion. The curves are the mean values over three training sessions with different seeds and the semi-transparent band represents the variance.}
	\label{fig:qfilter-num} 
	\vspace{5mm}
\end{figure}

\subsection{Extrapolation Error Reduction with Reward Filter}
During the training process,if SILP+ is steadily improving, the number of actions in the demonstration buffer that perform better than the policy should decrease. The comparison between the demonstrations and policy was done with a Q filter as used in \cite{8463162} and \cite{9561411}. The results were embedded in a behavior cloning loss and contributed to update the RL gradients. However, Q filter can also trigger extrapolation error and result in unstable and unreliable training. In this part, we did an ablation experiment to investigate how extrapolation error occurred in the Q filter and affected the training performance, and how the proposed reward based filter can alleviate the impact.  

In the experiments, we selected SAC-SILP+ as the algorithm with reward filter and replaced the reward filter with Q filter to form another comparison algorithm, called SAC-SILP+ with Q filter. First, we recorded the number of actions in the demonstration buffer that performed better than the actions from the policy during the training in Fig. \ref{fig:qfilter-num}. In the figure, the magenta curve and blue curve represent SAC-SILP+ with Q filter and SILP+ with our reward filter, respectively. From the curves, one can see that using the Q filter yields a large variance over the whole training process. In addition, the comparison was unstable before epoch 100, as one can observe from the downward and upward variations in the curve. In contrast, SILP+ with our reward filter (blue line) has a much more stable curve with lower variance. Furthermore, we notice that in our method, the number of actions in demonstrations that perform better than the policy is smaller than the one with Q filter after around epoch 50, which means the policy would need to imitate less from the expert while depending more on itself in SILP+. The results shown in Table \ref{tab:rfilters} indicate that the extrapolation error in the Q value not only results in unstable training but also deteriorates the success rate and training efficiency.
\begin{table}[!ht]
\renewcommand\arraystretch{1.2}
\small
	\caption{Sucesss rate and training time for two filter types (simulation)}
	\label{tab:rfilters}       
	\centering
\begin{tabular}{llll}
	\specialrule{0.1em}{0pt}{3pt}
	Performance           &  SILP+-Qfilter    & SILP+  \\
	\specialrule{0.1em}{3pt}{3pt}
	SR                 &  0.966           & 0.973 \\
	Time (h.m.s.)              &  06.07.47         & 5.22.58 \\
    \specialrule{0.1em}{3pt}{0pt}
\end{tabular}
\vspace{0mm}
\end{table}

\subsection{Actual robot experiments}

\begin{figure}[tb]
	\centering
	\begin{subfigure}[t]{0.4\textwidth}  
		\centering 
		\includegraphics[width=\textwidth]{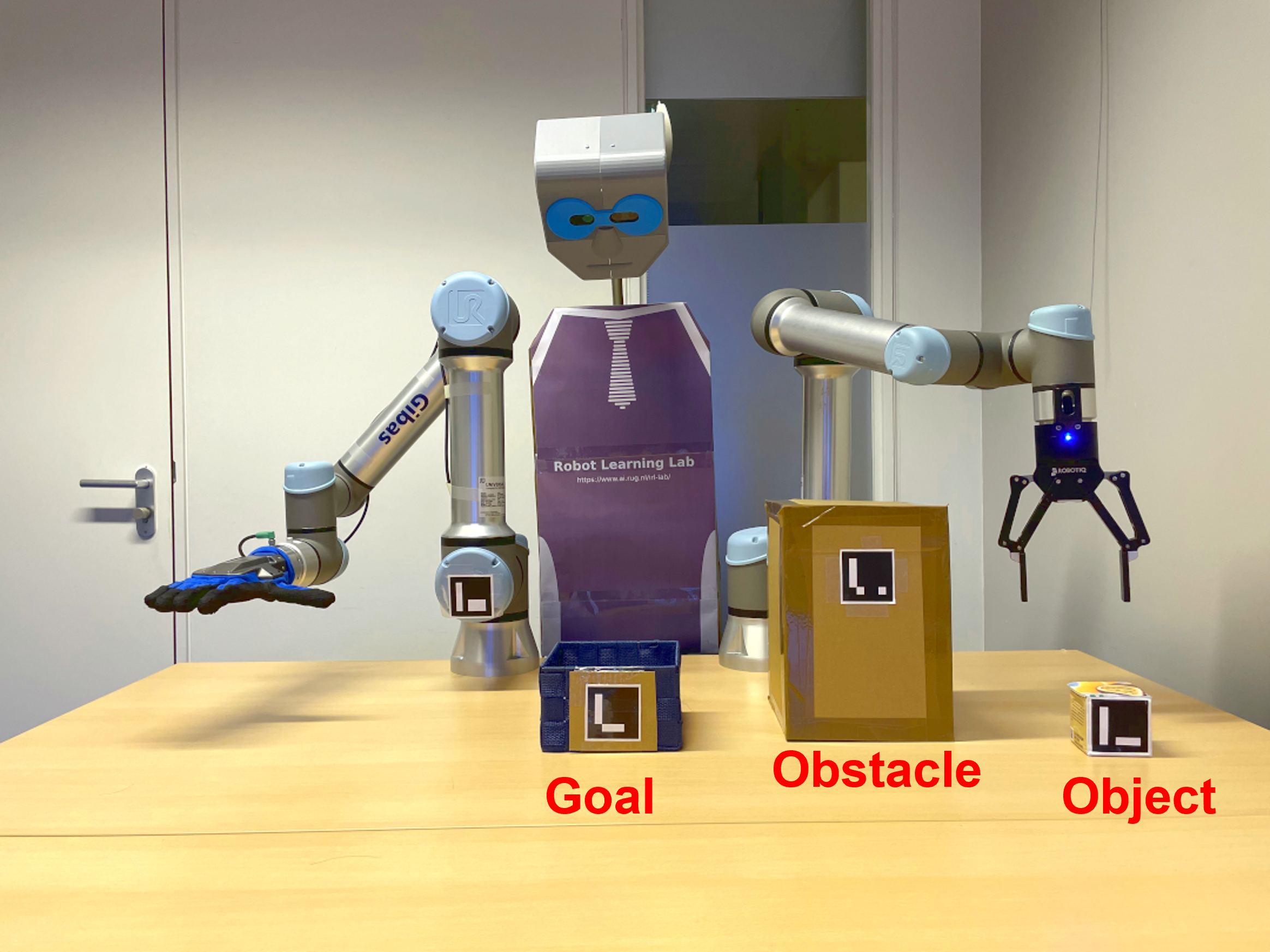}
		\caption{Initial state}
	\end{subfigure}
	\quad
	\begin{subfigure}[t]{0.4\textwidth}  
		\centering 
		\includegraphics[width=\textwidth]{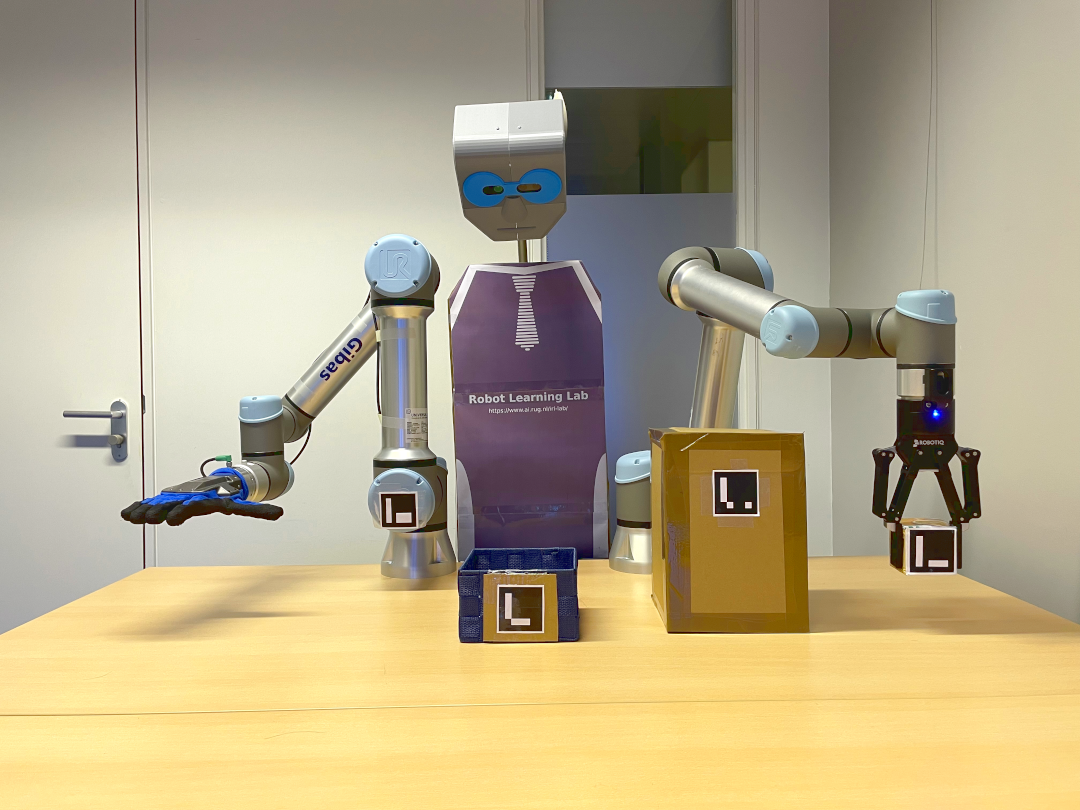}
		\caption{Grasp the object}
	\end{subfigure}
	\hspace{.01in}
	\begin{subfigure}[t]{0.4\textwidth}  
		\centering 
		\includegraphics[width=\textwidth]{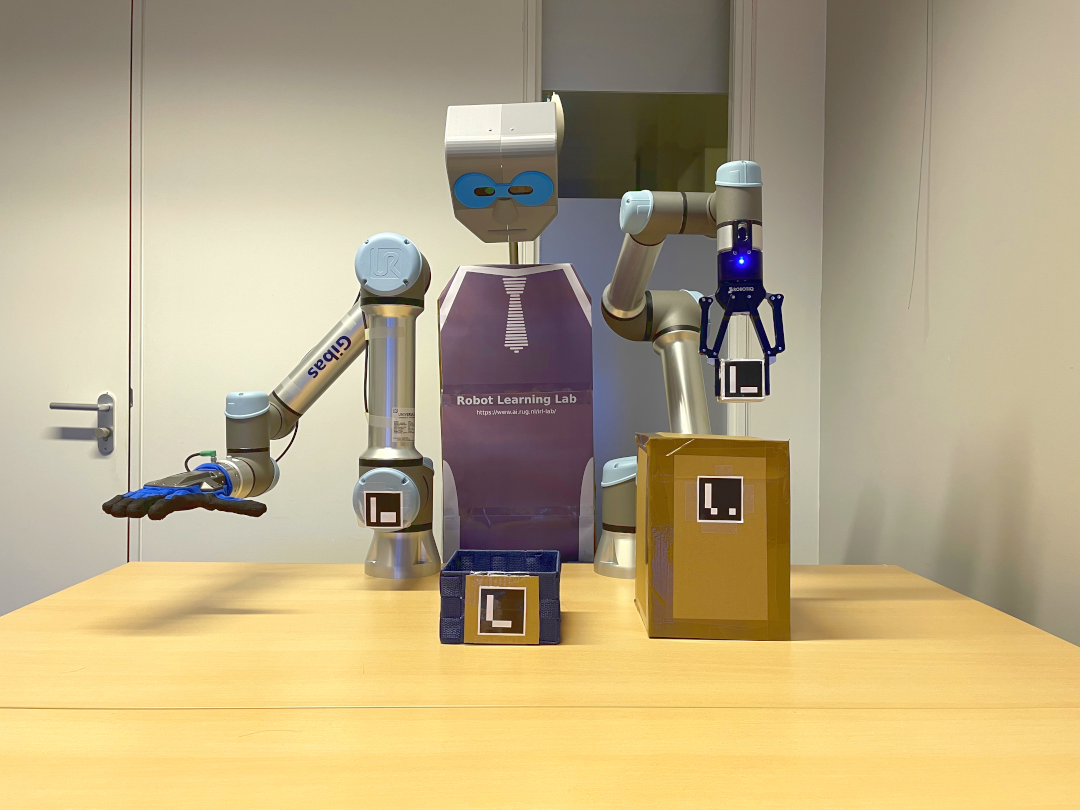}
		\caption{Move towards the goal}
	\end{subfigure}
	\quad
	\begin{subfigure}[t]{0.4\textwidth}  
		\centering 
		\includegraphics[width=\textwidth]{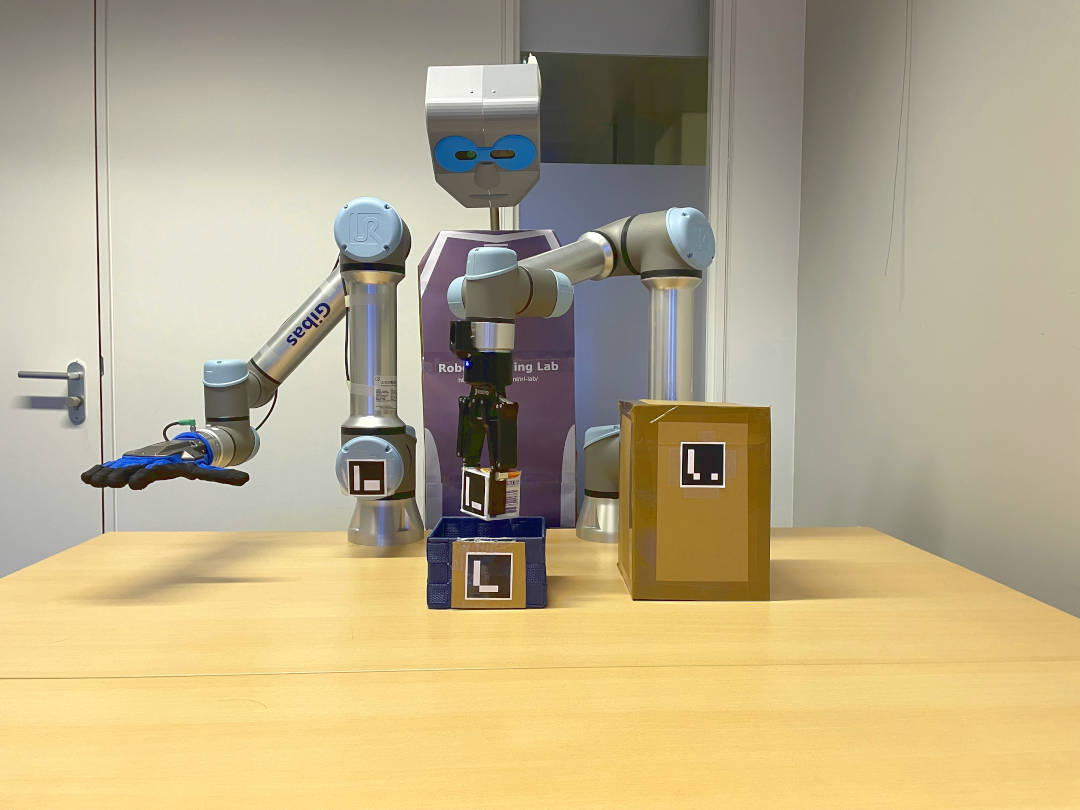}
		\caption{Place the object}
	\end{subfigure}
	\caption[physical experiments]
	{\small Pick and place task with physical robot and objects. The black and white square makers attached to the robot, objects are being used for the perception system for localization.} 
	\label{fig:illustration-real}
	\vspace{4mm}
\end{figure}

We designed several experiments on a physical UR5e robot mounted on a table (Universal Robots, S/N 20195501237).
First, we compare the performance of the best-performing policy SAC-SILP+ in simulation and on the real robot. Then we perform an additional comparison between SAC-SILP+ and the next-best contender, SAC-Demon on the real robot.
In the SAC-SILP+ test we used three different conditions in order to check the effectiveness of the simulation-trained policy being deployed on a pick-and place task in a physical context. The task is shown in Fig. \ref{fig:illustration-real}, in which three objects are randomly put on the table with the considerations of not arranging them far away from the trained workspace. The big cardboard box (0.2m $\times$ 0.2m $\times$ 0.3m) represents the obstacle that the robot should avoid during the movement; the small cube (0.07m $\times$ 0.07m $\times$ 0.07m) is the object that needs to be picked and placed in the blue basket (0.15m $\times$ 0.15m $\times$ 0.1m). The blue basket is the goal, functioning as a container for the cube. The pick-and-place task consist of the following four steps:
\begin{itemize}
    \item Objects recognition and localization: The perception system is supported by the Aruco marker method \cite{romero2018speeded} \cite{garrido2016generation}, which recognizes the markers and calculates the pose of the markers relative to the camera. Based on the poses of these markers, we can compute the objects’ locations relative to the base of the robot arm
    \item Pick up: The robot goes to the position of the cube with the gripper’s center located 20cm higher than the center of the cube in the Z direction; then the robot goes 10cm down to prepare for the grasping action and locate the cube inside the gripper; finally, the gripper closes to half-way to grasp the cube and goes 10cm up to pick the cube up
    \item Goal reaching:  The learned policy guides the robot to move towards the goal without colliding with the obstacle
    \item Place the object: Once the goal has been successfully reached, the gripper drops the cube into the basket. The process terminates if there is a collision or the policy runs out of pre-defined steps.
\end{itemize}
Note that in simulation training, the required positional accuracy for the successful reaching is 0.05m. Here, we use the reaching algorithm in a pick-and-place task. Therefore, we choose a larger-than 0.05m basket as the goal indicator taking into account the cube's size. Still, the required accuracy is nearly the same as in the simulation.

\begin{figure}
\centering
	\includegraphics[scale=0.25]{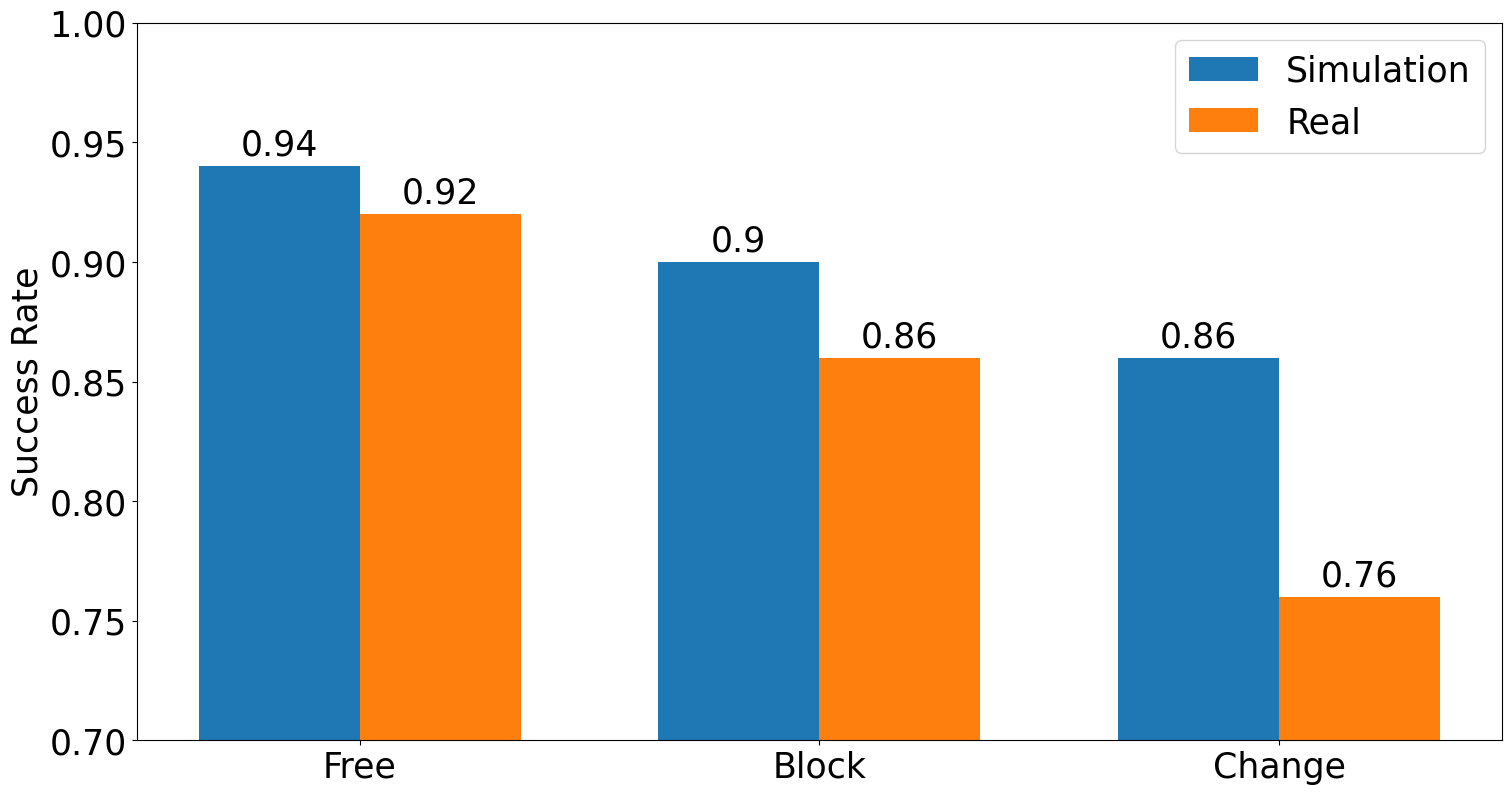}
	\caption{Success rates when SAC-SILP+ is applied on a real UR5e robot compared to the simulation environments. Please note that the origin of the Y-axis is not zero but 0.70.}
	\label{fig:real-sim-sr} 
	\vspace{5mm}
\end{figure}

The initial states of the robot, obstacle, cube and basket will affect the task's difficulty level and further affect the policy's performance. To fully test the performance of the algorithm on the real robot, we categorized the test scenarios into three types: Free (easiest), Block (normal) and Change (difficult). In the Free task, the obstacle is not blocking the path from the to-be-grasped object to the basket and the robot can move straight to the goal without considering the obstacle. Block means the obstacle locates between the robot's initial pose and the goal and the obstacle is a barrier for the robot's motion. The difficult one is Change, in which instead of a fixed goal and a fixed obstacle, the goal and/or obstacle will move to other locations during the execution of the policy. The difficulty level increases from Free to Block and then to Change. We first tested each scenario for 50 trials in the real environment and saved the trajectories of the end-effector and the perceived poses of the goal, obstacle and objects. Then, we repeated the real-world 
experiments in Gazebo using the same trained policy.

For the comparison between SAC-SILP+ and the next-best contender (SAC-Demon), we used 10 different scenarios with different obstacles, goals, and initial robot poses. The experimental setting is mostly the same as in the first experiment, except that this experiment focus on a reaching task instead of a pick-and-place task. We run the two simulator-trained policies for ten trials in each scenario and record the successful times. At the end of the testing, each policy will be tested 100 times.

\subsection{Actual UR5e robot - experimental results}

For the first SILP+ experiment, the success rates in physical and simulated environments were compared in Fig. \ref{fig:real-sim-sr}. 
The success rates decrease as the difficulty levels increase in simulation and physical environments, and the Sim2Real gap becomes more noticeable in more difficult scenarios. For the Free scenario, there were three and four failure trials in simulation and the real test, respectively. The three failures in both simulation and real trials were due to the fact that the obstacle's position located out of the trained workspace. Another failure in the real environment was a collision with the basket, which is attributed to the different experimental settings in simulation and the actual robot experiment. The policy was trained to reach a point in the simulation, but we applied it in a pick-and-place application in the physical world. There were four and seven failure cases in simulation and the real test in the Block scenario, respectively. Among the four common failures, three of them are due to the fact that the obstacle and/or goal were put out of the trained workspace; another one could be the algorithm's ability. There were three more failures in real experiments than in simulations. Two of them were due to collision and one was because of the out-of-basket dropping. 

Possible reasons include the measure error from the perception system and the different task settings between the simulation and physical experiment. In the Change situation, five failures were found in both simulation and real experiments. Interestingly, there were two failures in simulation that were not the case in the real test due to the complexity of the dynamic environment. The simulation was designed to replicate the experiments in the real-robot test. However, the time and speed of the goal and/or obstacle's change were hard to be replicated, which caused delays and collisions in simulation. In addition, there were seven failures in the physical experiments that did not happen in simulations; most of them were due to the measurement errors in the perception system. Generally, the causes of failure include locations out of familiar space, measurement errors, algorithm performance and time delay in the simulation. The external factors (e.g., environmental setting and sensor noises) played the most important role. 

For the second experiment, we summarize the results in Table \ref{tab:Sim2Real}.
\begin{table}[!ht]
\centering
\renewcommand\arraystretch{1.2}
\small
	\caption{Sim2Real success rate for SAC-SILP+ and SAC-Demon in a reaching task}
	\label{tab:Sim2Real}       
	\centering
\begin{tabular}{llll}
	\specialrule{0.1em}{0pt}{3pt}
	Method           &  SAC-SILP+    & SAC-Demon  \\
	\specialrule{0.1em}{3pt}{3pt}
	Simulation             &  0.90         & 0.87 \\
	Real world             &  0.90         & 0.75 \\
    \specialrule{0.1em}{3pt}{0pt}
\end{tabular}
\vspace{4mm}
\end{table}

The comparison between our proposed method and the next-best contender on 100 additional trials in 10 scenarios yielded a success rate of 0.90 for SAC-SILP+ and 0.75 for SAC-Demon (significant at $p < 0.001)$ on the real robot. In the simulated version of the test, the difference between SAC-SILP+ and SAC-Demon is small (0.90 vs 0.87) when compare to the results in Table \ref{tab:bl} which are based on 1000 trials. Failures, e.g., in 'scenario 5' are characterized by a discrepancy between training and testing conditions.

Based on these extensive empirical analyses, we conclude that the results of SAC-SILP+ are good. The small Sim2Real performance differences are expected to be solvable with better-matched experimental settings and improved perception systems. 

The edited videos for the two actual robot experiments can be accessed through \footnote{See \url{https://youtu.be/6Lgj_fVaCvo}} \footnote{See \url{https://youtu.be/zxjxQImLhxc}.}, respectively.

\section{Conclusion}\label{Conclusion}
In this article, we have proposed SILP+ to relieve humans from collecting diverse demonstrations in goal-conditioned motion planning tasks. With the guidance of self-imitation learning that utilizes the demonstrations from planning on past experience, we train a neural motion planner that generalizes substantially better than those learned directly from off-policy RL, behavior cloning or hind-sight experience replay. The experimental results show that the methods of dealing with collision failures significantly determine the performance. Both positive and negative guidance can boost performance, especially the positive demonstrations. However, hesitating and/or cyclic moves around obstacles will deteriorate the training process by confusing the agent, since pertinent information is lacking. Besides, we verified that the proposed Gaussian-process-based exploration near obstacles could accelerate the training by reducing unnecessary collisions, resulting in shortened training time. Furthermore, we analyzed the extrapolation error in the actor-critic neural networks and found that the extrapolation error would lead to unstable training and affect the learning efficiency and success rate.  This problem was solved by using a dedicated reward filter to obtain improved results.

We have explored a new way of embedding planning in the learning framework, while not adding much extra computation burden on the training process. The principle behind it may inspire the motion planning and reinforcement learning communities to design robust and efficient NMPs. In addition, the analysis and discussions on the collision solutions and extrapolation error reduction method could enhance studies related to safe and stable RL. The beneficial effect of the SILP+ method was confirmed for two common RL frameworks in current use,
i.e., SAC~\cite{haarnoja2018soft} and DDPG~\cite{lillicrap2015continuous}, with a clear advantage for the SAC-SILP+ combination.

Although we have tested SILP+ with a position controller on a UR5e robot arm, SILP+ can also be used with other controllers and other robotic platforms if an action model can be obtained to extract the MDP format demonstrations. Since the planning and learning modules in SILP+ are closely intertwined, there is a strong mutual influence. For example, a better exploration technique in RL will not only benefit RL but also generate better nodes for planning and further improve the quality of the behavior demonstrations. In future work, we will investigate how other planning techniques, such as MPCs \cite{9385847} \cite{9561298}, could guide the exploration to critical regions in the RL training process and generate more informative experiences for further demonstration planning in SILP+.

\appendix 
\section{Planning Distance Selection}\label{appendixA}
\renewcommand{\thetable}{A.\arabic{table}}
\setcounter{table}{0}
\renewcommand{\thefigure}{A.\arabic{figure}}
\setcounter{figure}{0}
In PRM, the choice of the function that is used to select the neighbors for local planning can affect the planning performance and the path quality \cite{amato1998choosing} \cite{kuffner2004effective}.
As aforementioned, we use the Euclidean distance as the metric to choose the neighbors to expand the path, but how to define the distance threshold $d$ is nontrivial. The best distance should be able to exclude unnecessary neighbor nodes that distant from the current node and discard neighbors that are too nearby to improve the planning efficiency by decrease the times of collision checking. 

We did an empirical decision-making experiment to explore the suitable distance. The distance space is defined as a discrete set: $\{0.1, 0.15, 0.2, 0.25,$ $ 0.3, 0.35, 0.4\}m$. First, we compare the planning success rate, planned path steps, and the planning time for the successful paths. Then, we combine the selected distance with SILP+ to train the policy and test the final success rate. We summarize the results in Table. \ref{tab:prm-dis}, from which we can see that the planning success rate increases when the neighbor distance $d$ increases. This trend is reasonable as the nodes in the roadmap are sparse since they are the explored states in one episode. So, a smaller distance will reduce the number of neighbor nodes within the distance, and the success rate decreases with fewer nodes within the defined $d$. The average planned path steps and planning time on each successfully planned path follow the decreasing trend with the distance increases. At last, we embed these different distances into SILP+ to compare the final task success rates and illustrate the result in the last row in Table.\ref{tab:prm-dis}. We can see that the distance of $0.15m$ gained the highest task success rate. The PRM-based path planning algorithm is utilized to provide demonstrations for SILP+, in which the task success rate and training efficiency are the most important metrics. So we choose the final distance of $0.15m$ as the final distance threshold in the experience-based planning module. 

\begin{table}[!htb]
	\small
	\centering
	\caption{The performance for different planning distance $d$}
	\label{tab:prm-dis}       
\begin{tabular}{llllllll}
	\hline\noalign{\smallskip}
	distance $d$ (m)    & 0.1        &  0.15      &    0.2      &  0.25      &    0.3      &  0.35      &    0.4      \\
	\noalign{\smallskip}\hline\noalign{\smallskip}
	SR(Planning)      & 0.8        &  0.8786    &  0.9028     & 0.9144     &  0.9214     & 0.9238     &   0.924    \\
    Avg Steps         & 6.39       &  5.614     &  5.064      & 4.676      &  4.402      & 4.152      &   3.920     \\
	Planning Time              & 0.242      &  0.218     &  0.202      & 0.191      &  0.184      & 0.177      &   0.171     \\
	SR(SILP+)          & 0.961      &  \textbf{0.965}     &  0.933      & 0.828      &  0.782      & 0.745      &   0.663     \\
	\noalign{\smallskip}\hline
\end{tabular}
\vspace{0mm}
\end{table}

\section{Collision Model}\label{appendixB}
\renewcommand{\thetable}{B.\arabic{table}}
\setcounter{table}{0}
\renewcommand{\thefigure}{B.\arabic{figure}}
\setcounter{figure}{0}

Collision checking is one of the most computationally expensive modules in path planning \cite{canny1988complexity}. Researchers tried to accelerate the collision checking process with neural networks. For instance, Kew et al. \cite{chase2020neural} proposed a neural network called ClearanceNet to predict the minimal distance between the robot and the workspace and use this prediction to infer the collisions. L{\"u}tjens et al. \cite{8793611} predicted collision probability based on a set of LSTM networks. Tuan et al. \cite{tran2020predicting} utilized Contractive AutoEncoder (CAE) to learn the latent representation of the collision-free space in order to predict the validation of robot configurations.

\begin{figure}[!ht]
\centering
	\includegraphics[scale=0.6]{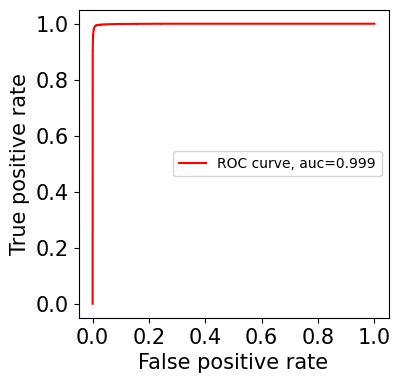}
	\caption{ROC curve of the collision checking model}
	\label{fig:roc} 
	\vspace{2mm}
\end{figure}

In this work, we learned a neural network model $\mathcal{M}$ that predicts the collision probability given the robot's joints configuration and obstacle's information. The model has three hidden layers (512, 256 and 64 nodes each) and an output for the prediction of collision probability. The dataset contains 485$k$ configurations and among them, $90\%$ and $10\%$ were used as the training and testing dataset. The evaluated accuracy, recall, precision and specificity are 0.991, 0.984, 0.989 and 0.995, respectively, under the discrimination threshold of 0.5. The ROC curve is illustrated in Fig. \ref{fig:roc}. Ultimately, the output from $\mathcal{M}$ can be used to predict the collision probability between two states to improve the planning efficiency and the quality of the planned demonstrations.

\section*{Acknowledgements} 
We want to thank the Reviewers for taking the time and effort necessary to review the manuscript. We sincerely appreciate all valuable comments and suggestions, which helped us improve the manuscript's quality. In addition, we are grateful to Dr.Hamidreza Kasaei for his helpful discussions and experimental contributions. We are also thankful to Weijia Yao for the early discussions on problem formulation and article proofreading. Finally, we thank the Center for Information Technology of the University of Groningen for their support and for providing access to the Peregrine high-performance computing cluster.

\bibliography{ref}
\bibliographystyle{elsarticle-num}
\end{document}